\documentclass{article} 
\usepackage{iclr2027_conference,times}

\usepackage{amsmath,amsfonts,bm}

\def\eqref#1{equation~\ref{#1}}

\def\1{\bm{1}}

\DeclareMathAlphabet{\mathsfit}{\encodingdefault}{\sfdefault}{m}{sl}
\SetMathAlphabet{\mathsfit}{bold}{\encodingdefault}{\sfdefault}{bx}{n}

\usepackage{hyperref}
\usepackage{url}
\usepackage{enumitem}
\usepackage{caption}
\usepackage{microtype}
\usepackage[T1]{fontenc}
\usepackage{booktabs}
\usepackage{graphicx}
\usepackage{pifont}
\usepackage[dvipsnames]{xcolor}
\usepackage[most]{tcolorbox}
\usepackage{colortbl}
\usepackage{tabularx}
\usepackage{wrapfig}
\usepackage[table]{xcolor}
\usepackage{pgf}
\usepackage{subcaption}
\usepackage{longtable}
\usepackage{array}
\usepackage{multirow}

\newcommand{\cmark}{\textcolor{ForestGreen}{\ding{51}}}
\newcommand{\xmark}{\textcolor{red}{\ding{55}}}
\definecolor{my_yellow}{RGB}{255,244,169}
\definecolor{my_dark_yellow}{RGB}{255,225,159}
\definecolor{my_blue}{RGB}{217,241,254}
\definecolor{my_green}{RGB}{227,250,234}
\definecolor{my_purple}{RGB}{243,227,252}

\newcommand{\gaincell}[3]{%
    \pgfmathparse{sqrt(max(0,min(1,(#1-#2)/#3)))}%
    \let\heatvalue\pgfmathresult%
    \pgfmathparse{%
        ifthenelse(\heatvalue<0.5,
        200*\heatvalue,
        200*\heatvalue-100)%
    }%
    \pgfmathtruncatemacro{\heatpercent}{\pgfmathresult}%
    \ifdim\heatvalue pt<0.5pt
        \edef\heatcolor{my_yellow!\heatpercent!white}%
    \else
        \edef\heatcolor{my_dark_yellow!\heatpercent!my_yellow}%
    \fi
    \expandafter\cellcolor\expandafter{\heatcolor}#1%
}

\title{Behavior-Grounded Semantic Enrichment for Financial Fraud Modeling and Reasoning}

\author{
Linbo Shao\footnotemark[1], 
Huilin He\footnotemark[1], 
Yating Lou, 
Dawei Cheng\footnotemark[2]\\
School of Computer Science and Technology, Tongji University, Shanghai, China\\
\texttt{\{2634102, huilin3, 2612120, dcheng\}@tongji.edu.cn}\\
\small $^{*}$Equal contribution \quad $^{\dagger}$Corresponding author
}

\iclrfinalcopy 
\begin{document}

\maketitle

\begin{abstract}
In financial fraud detection, rich semantic context can provide important evidence for transaction behavior modeling and fraud reasoning. However, public real-world financial datasets often lack rich semantics due to privacy constraints. Consequently, synthetic datasets incorporate generated semantics, but at the cost of behavioral realism; textual descriptions for contextual reasoning remain scarce. We address this gap through a semantic enrichment framework grounded in original transaction behavior to simulate multimodal financial data. We (1) propose a multi-agent semantic enrichment framework that generates interpretable financial semantics grounded in transaction behavior through role-specialized agents and consistency refinement, and (2) newly contribute a valuable multimodal financial fraud dataset, MS-FFSD, enriched with structured semantics and textual semantics while preserving real-data-grounded transaction behavior. Furthermore, we systematically analyze the quality and utility of semantic enrichment. Results demonstrate statistical fidelity and framework generalizability, while showing that richer semantics benefit fraud modeling and context-aware LLM reasoning. Overall, this work advances multimodal financial fraud research and bridges emerging LLM and multi-agent capabilities with operational anti-fraud practice. The framework and dataset are released at \url{https://github.com/AI4Risk/MS-FFSD}.
\end{abstract}

\section{Introduction}

Fraud detection is a critical task across diverse real-world domains~\citep{ref2,ref51}, and corresponding public datasets range from social reviews~\citep{ref34,ref35} to consumer finance~\citep{ref36,ref5}, with varying modality richness. Among these modalities, semantic and textual context can provide important evidence for understanding fraudulent behavior, and recent advances in large language models (LLMs) have created new opportunities for context-aware fraud analysis~\citep{ref38,ref39,ref50}. However, financial fraud detection presents a distinct challenge: transaction data constitute a central resource across banking~\citep{ref7} and e-commerce systems~\citep{ref6}, yet public financial transaction datasets remain semantically sparse~\citep{ref3,ref33}, thereby limiting support for leveraging semantics in fraud modeling and reasoning.

In real-world financial systems, transaction data are often accompanied by rich semantic information describing participating entities and transaction scenarios, which can support risk modeling and reasoning~\citep{ref47,ref46,ref49}. Yet upon public release, privacy constraints necessitate the removal or anonymization of semantically rich information, leaving the resulting datasets dominated by numerical features and coded attributes~\citep{ref3,ref45,ref48}. This semantic scarcity has motivated the development of synthetic financial datasets that provide richer information, such as user profiles, merchant categories, and geographic attributes~\citep{ref4,ref32}. However, because their transaction behaviors are largely generated through hand-crafted rules or simulators, these datasets inevitably differ from real financial systems in structural distributions and behavioral complexity~\citep{ref43,ref37}. 

This creates an important gap in existing financial transaction resources: public real-world datasets retain realistic behavioral structure but limited semantic context, whereas rule-driven synthetic datasets provide richer semantics but limited behavioral realism. Moreover, existing datasets rarely provide entity-level textual descriptions, limiting their support for semantic reasoning. Table~\ref{tab:dataset_comparison} summarizes these differences across representative public financial transaction datasets. This motivates behavior-grounded semantic enrichment. To model the multidimensional and interdependent financial semantics involved, we propose a multi-agent framework. Designed for transaction data with user--merchant interactions and known feature meanings, the framework integrates observed transaction behavior with external statistical and domain knowledge to guide structured and textual semantic generation, while preserving the original transaction structure and fraud labels. Specifically, the framework comprises agents for temporal construction, user semantic initialization, merchant semantic assignment, cross-entity consistency refinement, and entity-level textual description generation. The agents share transaction and entity states, with generation guided by external knowledge, global constraints, and fraud-aware preservation. The resulting structured semantics augment the tabular modality, while textual semantics form the textual modality. Together, they yield multimodal financial transaction data that support both fraud modeling and contextual reasoning.

As a semantically rich and publicly accessible data contribution, we construct and release MS-FFSD, a multimodal financial fraud dataset supporting both fraud modeling and semantic reasoning. S-FFSD~\citep{ref11} provides a suitable foundation because it preserves feature meanings and explicit user--merchant interactions, while its transactions were generated by a graph-based model trained on real-world data~\citep{ref12}, rather than through hand-crafted rules. Applying the framework, we enrich the original transactions with structured and textual semantics while preserving their real-data-grounded behavior. We further apply the framework to two private real-world financial datasets to establish its broader applicability across financial datasets.

\begin{table*}[t]
\centering
\caption{
Comparison of public financial transaction datasets. Real-Data Grounded denotes real transaction origin or generation from real transaction data; Feature Interpretability and Original Representation indicate retained feature meanings and untransformed values; Transaction, Entity, and Textual indicate available semantic information. MS-FFSD covers all dimensions.}
\label{tab:dataset_comparison}
\renewcommand{\arraystretch}{1.5}
\setlength{\tabcolsep}{3pt}
\resizebox{\textwidth}{!}{
\begin{tabular}{lcccccc}
\toprule
& \textbf{Data Foundation}
& \multicolumn{2}{c}{\textbf{Feature Information}}
& \multicolumn{3}{c}{\textbf{Semantic Information}} \\
\cmidrule(lr){2-2}
\cmidrule(lr){3-4}
\cmidrule(lr){5-7}

\textbf{Dataset}
& \textbf{Real-Data Grounded}
& \textbf{Feature Interpretability}
& \textbf{Original Representation}
& \textbf{Transaction}
& \textbf{Entity}
& \textbf{Textual} \\
\midrule

Elliptic++~\citep{ref31}
& \cmark
& \cmark
& \cmark
& \cmark
& \cmark
& \xmark \\

FraudEcom~\citep{ref6}
& \cmark
& \cmark
& \cmark
& \cmark
& \xmark
& \xmark \\

IEEE-CIS~\citep{ref5}
& \cmark
& \xmark
& \cmark
& \xmark
& \xmark
& \xmark \\

Credit Card Fraud~\citep{ref7}
& \cmark
& \xmark
& \xmark
& \xmark
& \xmark
& \xmark \\

\midrule

Sparkov~\citep{ref8}
& \xmark
& \cmark
& \cmark
& \cmark
& \cmark
& \xmark \\

CCT~\citep{ref10}
& \xmark
& \cmark
& \cmark
& \cmark
& \cmark
& \xmark \\

SAML-D~\citep{ref32}
& \xmark
& \cmark
& \cmark
& \cmark
& \cmark
& \xmark \\

\midrule

S-FFSD~\citep{ref11}
& \cmark
& \cmark
& \cmark
& \xmark
& \xmark
& \xmark \\

\rowcolor{my_blue}
\textbf{MS-FFSD (ours)}
& \cmark
& \cmark
& \cmark
& \cmark
& \cmark
& \cmark \\

\bottomrule
\end{tabular}
}
\end{table*}

Our main contributions are as follows:
\begin{itemize}[leftmargin=15pt,topsep=0pt]
\setlength\itemsep{0em}

\item We propose a multi-agent semantic enrichment framework that generates financial semantics grounded in transaction behavior through role-specialized agents and consistency refinement.

\item We contribute and release MS-FFSD, a valuable multimodal financial fraud dataset enriched with structured and textual semantics while preserving real-data-grounded transaction behavior.

\item We systematically analyze the quality and utility of semantics for financial fraud analysis. Specifically, we assess statistical fidelity and framework generalizability, and provide insights into how semantic enrichment affects financial fraud modeling and semantic reasoning.

\end{itemize}

\section{Related Works}

\paragraph{Real-World Financial Transaction Datasets.}
Publicly available real-world financial transaction datasets preserve transaction records and fraud signals from operational systems, but privacy and regulatory constraints necessitate semantic information to be removed or anonymized before release~\citep{ref3,ref42,ref44}. Elliptic++~\citep{ref31} preserves real Bitcoin transactions and wallet entities, but the entities remain pseudonymous rather than real-world identities. FraudEcom~\citep{ref6} provides interpretable transaction context but limited entity semantics. IEEE-CIS~\citep{ref5} retains transaction and identity features, but many variables are anonymized into uninterpretable numerical values or categorical codes. Credit Card Fraud~\citep{ref7} further replaces original values with transformed representations. Consequently, these datasets support fraud modeling with real transaction behavior but provide limited semantics for LLM-based fraud detection and contextual reasoning.

\paragraph{Synthetic Financial Transaction Datasets.}
Synthetic datasets alleviate the above limitation by constructing entity and transaction semantics through simulation~\citep{ref37,ref4,ref43}. Sparkov~\citep{ref8} generates transactions from predefined user profiles and fraud patterns. CCT~\citep{ref10,ref9} further models user motivations and merchant dependencies to generate transaction sequences. SAML-D~\citep{ref32} extends this paradigm to anti-money laundering by jointly simulating transaction semantics and laundering patterns. In these datasets, transaction structures are therefore largely determined by predefined semantics and simulation mechanisms, inevitably differ from the behavioral complexity of real financial systems.

\paragraph{Semantic Enrichment Frameworks.}
The above limitations motivate behavior-grounded semantic enrichment. Sato~\citep{ref54} recovers column semantics from table context; ReTabAD~\citep{ref42} restores dataset, column and label semantics from original documentation. Yet for privacy reasons, financial semantics are irreversibly lost, motivating generation from observable evidence. In general, LogicNLG~\citep{ref55} and PLOG~\citep{ref56} generate logical text grounded in tabular facts. In the financial domain, semantic enrichment requires generating inferential text grounded in transaction behavior. Recent multi-agent frameworks support this process through agent coordination and iterative refinement~\citep{ref57,ref60,ref58,ref59}. Given the multidimensionality and interdependent nature of financial semantics, our framework assigns agents distinct semantic roles with dedicated external knowledge and priors, and enforces consistency both with observed transactions and across related entities.

\begin{figure}[t]
  \centering
  \includegraphics[width=1\linewidth]{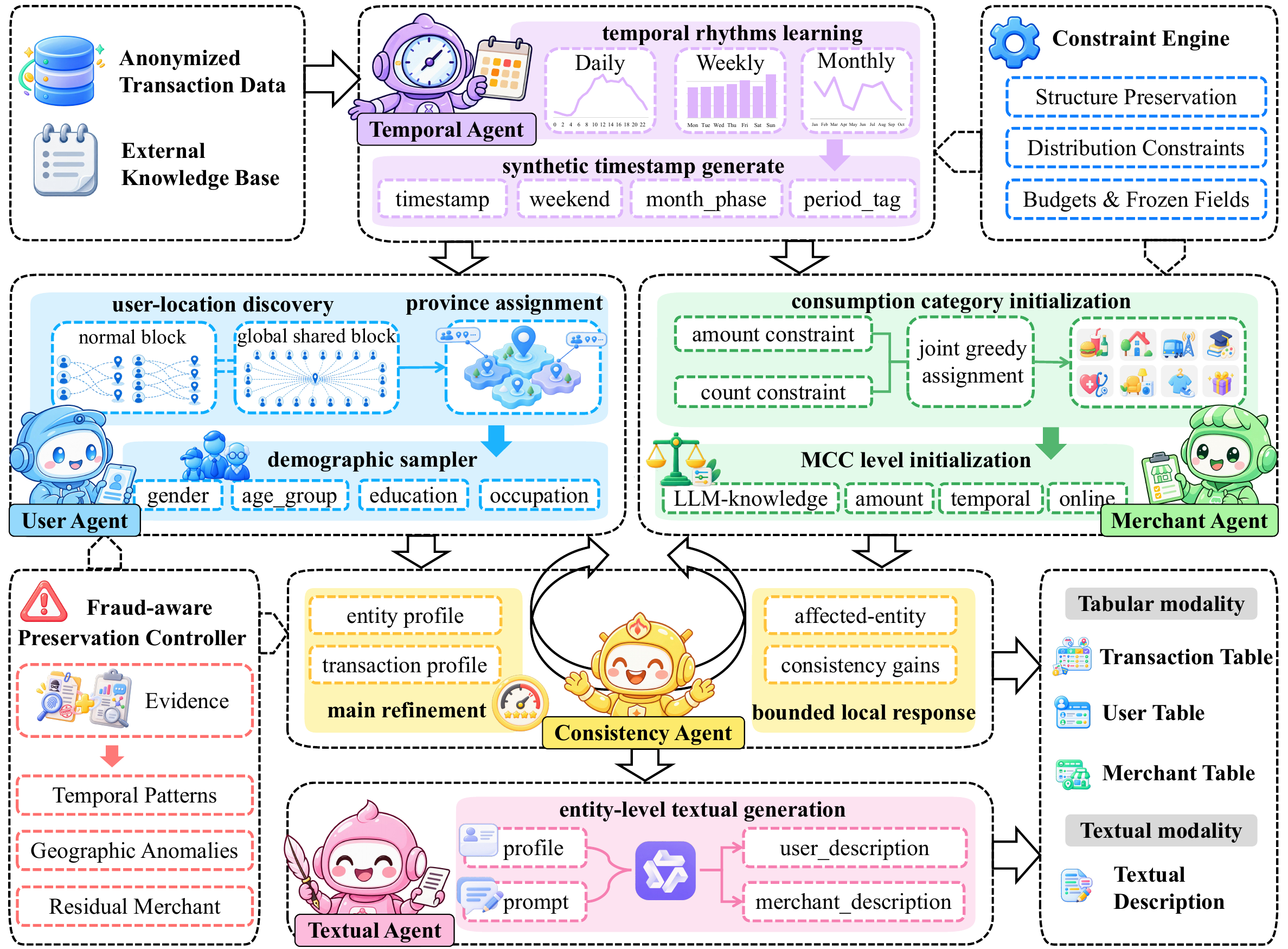}
  \caption{Overview of the proposed multi-agent semantic enrichment framework. The agents sequentially construct synthetic timestamp, user attributes, merchant semantics, cross-entity consistency, and entity-level textual descriptions, while shared knowledge, constraint, and fraud-preservation mechanisms, resulting in multimodal financial transaction data with tabular and textual modalities.}
  \label{fig:model}
\end{figure}

\section{Multi-Agent Semantic Enrichment Framework}

\subsection{Framework Overview}

We propose a multi-agent semantic enrichment framework grounded in transaction behavior, shown in Figure~\ref{fig:model}. Let $\mathcal D$ denote the original transactions and $\mathcal Z_S=(\boldsymbol{\tau},\boldsymbol r,\boldsymbol q)$ the temporal, user, and merchant semantic state. The structured enrichment is summarized by the following unified objective:

\begin{equation}
\mathcal F_S(\mathcal Z_S)
=
F_T(\boldsymbol{\tau})
+\lambda_U F_U(\boldsymbol r)
+\lambda_M F_M(\boldsymbol q;\boldsymbol r,\boldsymbol{\tau})
+\lambda_C F_C(\boldsymbol r,\boldsymbol q;\boldsymbol{\tau},\mathcal G_{UM}),
\label{eq:framework_objective}
\end{equation}

The four terms model temporal construction, user initialization, merchant initialization, and cross-entity refinement under the external knowledge base $\mathcal K$, feasible region $\mathcal C(\mathcal D,\mathcal K)$, and observed user--merchant graph $\mathcal G_{UM}$. They are realized through staged agent collaboration while preserving the original transaction structure and fraud labels. The Textual Agent then generates entity-level descriptions from the refined structured state. Detailed formulations are provided in Appendix~\ref{app:framework_details}.

\subsection{Structured Semantic Initialization}

Structured semantic enrichment grounds anonymous transactions in interpretable temporal, user, and merchant contexts. The three initialization agents are formulated under a unified objective:

\begin{equation}
F_A=\sum_{j\in\mathcal J_A}\lambda_j\mathcal L_j^A,\quad
A\in\{T,U,M\},\quad
\mathcal J_T=\{r,u,b\},\ 
\mathcal J_U=\{g_n,g_a,d\},\ 
\mathcal J_M=\{c,a,s\}.
\label{eq:initialization_objectives}
\end{equation}

For the Temporal Agent, $r$, $u$, and $b$ denote global rhythm matching, user-specific temporal patterns, and fraud-burst preservation, respectively. It reconstructs order-preserving timestamps, using IEEE-CIS\footnote{\url{https://www.kaggle.com/c/ieee-fraud-detection/}} as a public reference for coarse temporal rhythms and incorporating user-level patterns when sufficient evidence is available. For the User Agent, $g_n$ and $g_a$ capture geographic matching by user count and transaction amount, while $d$ represents conditional demographic distributions. The agent combines observed user--location interactions with external priors to initialize geographic attributes, gender, age group, education, and occupation. For the Merchant Agent, $c$, $a$, and $s$ represent merchant-count, transaction-amount, and semantic-compatibility terms. These objectives guide hierarchical category assignment by balancing distributions and semantic compatibility.

\subsection{Cross-Entity Semantic Refinement}

Independently initialized user and merchant semantics may remain inconsistent with observed cross-entity interactions. We introduce a Consistency Agent that iteratively refines user and merchant semantics over graph $\mathcal G_{UM}$. Its objective combines user and merchant side consistency:

\begin{equation}
F_C=
\mathcal L_U^C+\mathcal L_M^C,
\label{eq:consistency_objective}
\end{equation}

where the two terms capture cross-entity consistency. At iteration $t$, let $Z^{(t)}$ be the current state and $Z_{-e}^{(t)}$ exclude entity $e$. Each candidate update is evaluated by its net consistency gain:

\begin{equation}
\begin{aligned}
\Delta_e(z'_e;Z^{(t)})={}&
S_e(z'_e;Z_{-e}^{(t)})
-
S_e(z_e^{(t)};Z_{-e}^{(t)})
-\kappa_e\mathbb I[z'_e\neq z_e^{(t)}].
\end{aligned}
\label{eq:semantic_gain}
\end{equation}

where $\kappa_e$ penalizes unnecessary semantic changes. Strictly positive-gain updates satisfying distributional and modification constraints are accepted. Changes on one side activate only affected counterparts on the other, enabling bidirectional responses without global reassignment. Refinement terminates when no admissible update remains or the modification budgets are exhausted.

\subsection{Textual Semantic Generation}

After structured refinement, the Textual Agent converts the refined entity states into natural-language descriptions using Qwen3-32B~\citep{ref29}. The generation process is summarized as

\begin{equation}
\boldsymbol d^\star=A_X(\mathcal D,\mathcal Z_S^\star,\mathcal K),
\qquad
\mathcal Z^\star=(\mathcal Z_S^\star,\boldsymbol d^\star).
\label{eq:textual_generation}
\end{equation}

where $\boldsymbol d^\star$ denotes the generated entity descriptions and $\mathcal Z^\star$ the final enriched semantic state. Generation is grounded in the refined structured semantics and aggregated transaction evidence associated with each entity. Single-transaction users are described using deterministic templates due to insufficient evidence for reliable behavioral aggregation, while multi-transaction users use LLM-based generation from aggregated behavioral profiles. Merchant descriptions combine business semantics, transaction behavior, and associated user profiles, with category-level memory reducing repetitive concepts. The Textual Agent does not modify the refined structured state during generation. The complete generation prompts and deterministic template are provided in Appendix~\ref{app:text_generation_prompts}.

\subsection{Agent Collaboration and Shared Control}

The specialized agents operate over shared transaction and entity states, forming a staged workflow in which temporal semantics support entity initialization, user and merchant states are iteratively refined through observed interactions, and the resulting structured states are passed to the Textual Agent. Cross-entity updates are restricted to affected counterparts and bounded by shared constraints, enabling controlled information exchange without unconstrained propagation.

Agent collaboration is governed by three shared components. The External Knowledge Base $\mathcal K$ provides statistical and domain knowledge for semantic construction, with its sources, granularity, temporal alignment, and roles detailed in Appendix~\ref{app:external_knowledge}. The Constraint Engine defines the feasible region $\mathcal C(\mathcal D,\mathcal K)$ through structural preservation, distributional consistency, category restrictions, and modification budgets. The Fraud-aware Preservation Controller further protects evidence-supported fraud patterns from being removed during enrichment. Together, these components coordinate agent updates while preserving the structural, statistical, and fraud-related properties of the original data.

\begin{wraptable}{!h}{0.5\textwidth}
\centering
\caption{Summary statistics of MS-FFSD.}
\label{tab:sffsd_statistics}
\setlength{\tabcolsep}{10pt}
\small
\begin{tabular}{lr}
\toprule
\rowcolor{my_blue}
\textbf{Statistic} & \textbf{Value} \\
\midrule
Transactions & 77,881 \\
Users & 30,346 \\
Merchants & 886 \\
Normal transactions & 31.31\% \\
Fraud transactions & 6.75\% \\
Unlabeled transactions & 61.94\% \\
Observation period & Jan.--Oct. 2021 \\
\bottomrule
\end{tabular}
\end{wraptable}

\section{MS-FFSD Dataset}

\subsection{Dataset Overview}

MS-FFSD is the multimodal financial fraud dataset constructed and released in this work. It builds on S-FFSD~\citep{ref11}, whose transaction structure was generated by a graph-based generative model trained on real-world data, and enriches it with semantics grounded in observed transaction behavior and informed by real-world statistical distributions and domain knowledge. The enrichment process introduces two forms of semantic information, \textit{Structured Semantics} and \textit{Textual Semantics}, while preserving the original transaction records, interactions between users and merchants, and fraud labels. The resulting dataset integrates the original transaction data with enriched semantic information to support multimodal financial fraud research. Table~\ref{tab:sffsd_statistics} summarizes its scale, label, and period.

\begin{figure}[t]
  \centering
  \includegraphics[width=1\linewidth]{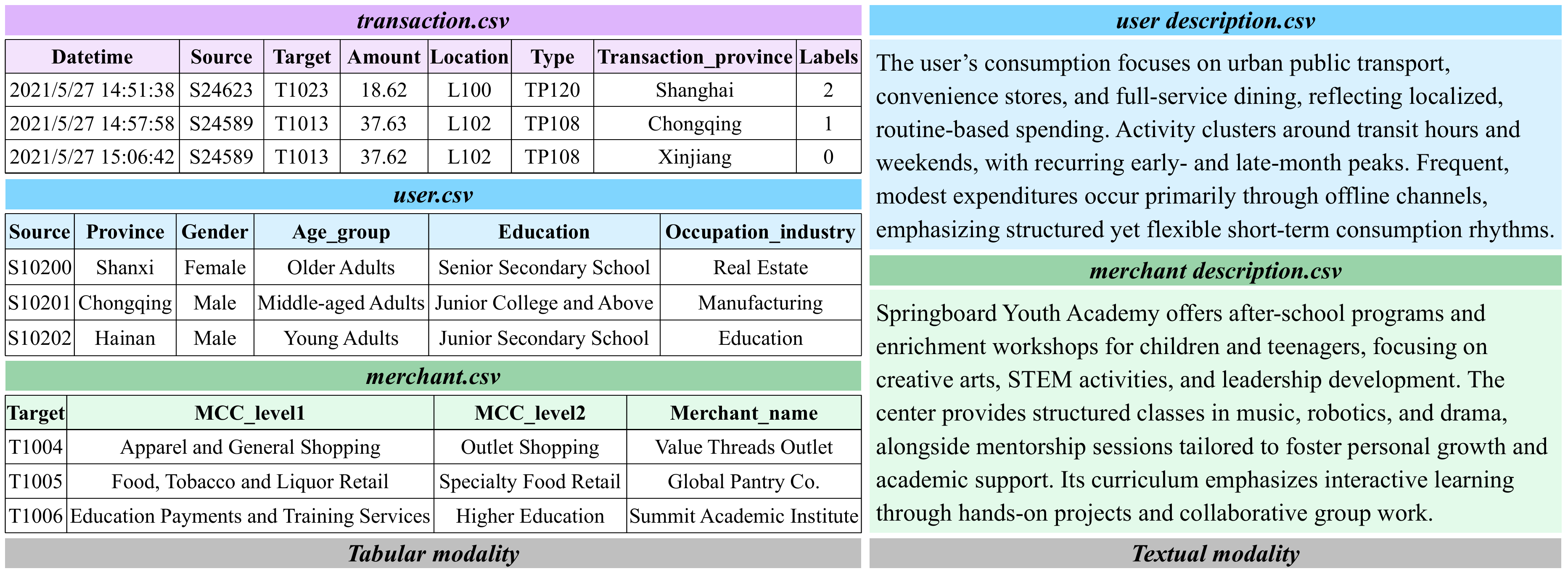}
  \caption{Data organization and modalities of MS-FFSD. Tabular modality: \textit{transaction} preserves original fields and adds temporal and geographic attributes. \textit{user} provides demographic and geographic attributes. \textit{merchant} provides hierarchical MCC categories and identities. Textual modality: \textit{description} summarizes user behavior and merchant business characteristics in natural language.}
  \label{fig:table}
\end{figure}

\subsection{Data Organization and Modalities}

MS-FFSD comprises two modalities: tabular and textual, as illustrated in Figure~\ref{fig:table}. The transaction table preserves all original transaction fields and augments with generated temporal and geography. The user table provides demographic and geographic attributes, including province, gender, age group, education, and occupation, while the merchant table provides hierarchical business categories and merchant names. These tables constitute the \textit{tabular modality}. The corresponding description files contain entity-level natural-language descriptions of users and merchants, representing behavioral characteristics and business activities, forming the \textit{textual modality}.

\begin{figure}[!b]
  \centering
  \includegraphics[width=1\linewidth]{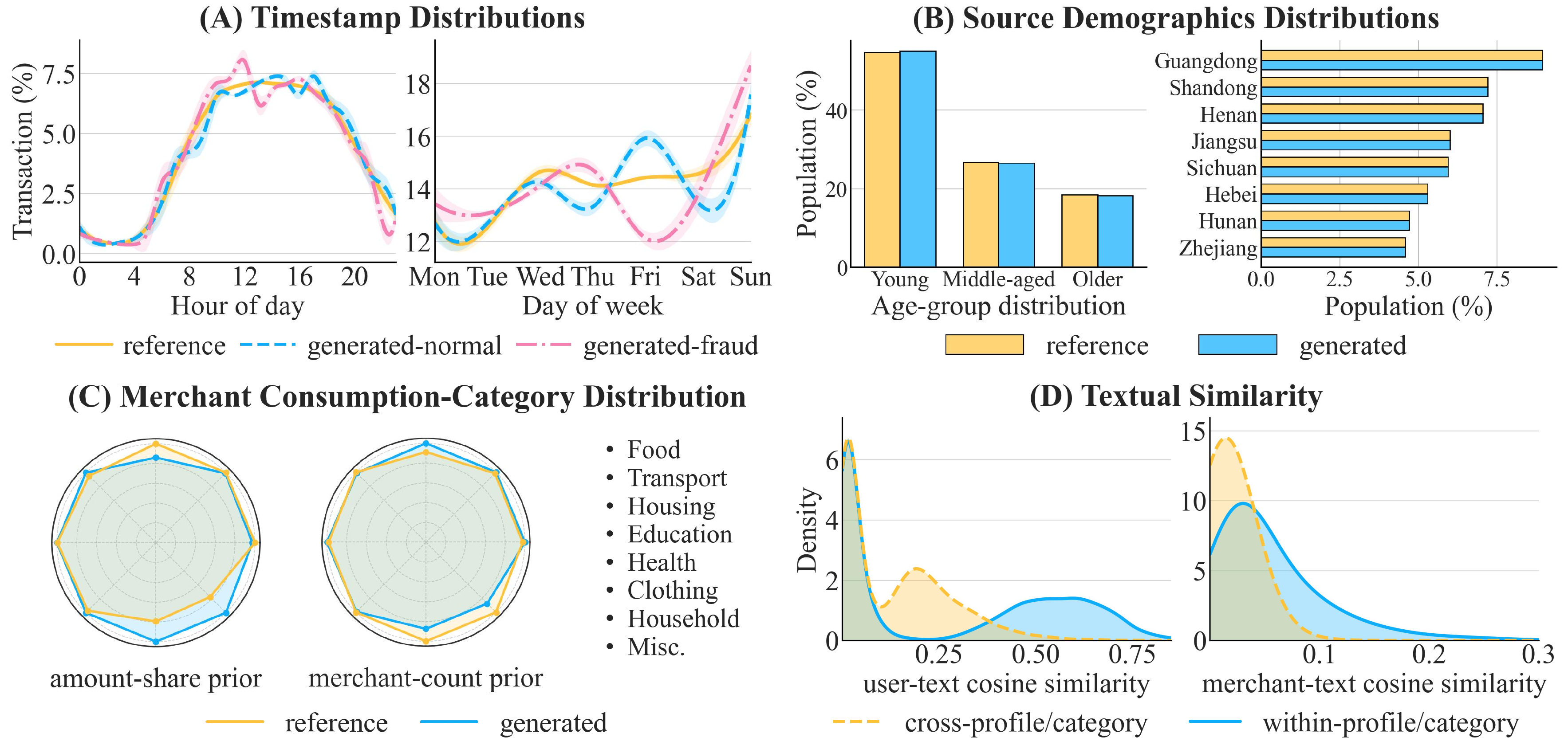}

  \caption{Semantic enrichment quality of MS-FFSD.
  (A) Generated timestamps follow reference rhythms.
  (B) Generated user demographics align with reference age-group and provincial distributions.
  (C) Merchant radar plots show category relative deviations from reference amount and count priors.
  (D) Textual descriptions show clear within-group coherence and cross-group discriminability.}
  \label{fig:semantic_quality}

  \begin{subcaptiongroup}
    \phantomsubcaption\label{fig:semantic_quality_time}
    \phantomsubcaption\label{fig:semantic_quality_user}
    \phantomsubcaption\label{fig:semantic_quality_merchant}
    \phantomsubcaption\label{fig:semantic_quality_text}
  \end{subcaptiongroup}
\end{figure}

The enriched semantics can be viewed from three semantic perspectives. \textit{Transaction-level Structured Semantics} characterize individual transactions through temporal and geographic context in addition to the original transaction attributes, including amounts and types. \textit{Entity-level Structured Semantics} describe users and merchants through demographic, geographic, and business-category attributes. \textit{Textual Semantics} further capture user behavioral characteristics and merchant identities and business activities through natural-language descriptions. Single-transaction users are described using deterministic templates, while multi-transaction users receive LLM-generated descriptions based on aggregated behavioral patterns; merchant descriptions are generated from category semantics, transaction behavior, and associated user profiles. Together, these three perspectives provide a comprehensive semantic characterization of financial transactions and entities. Appendix~\ref{app:dataset_details} provides complete field definitions, user semantic attributes, merchant category definitions, and textual description statistics.

\section{Evaluation}

We first assess whether the generated semantics exhibit distributional fidelity and semantic informativeness, establishing the validity of the enrichment results. We then investigate how progressively enriched semantic information changes the way models exploit and interpret financial transaction behavior. From the predictive perspective, we evaluate its value through graph-based fraud detection and further examine the broader applicability of the enrichment framework on two additional private real-world datasets. From the reasoning perspective, we analyze how progressively enriched semantics support LLM-based contextual interpretation of transaction behavior.

\subsection{Semantic Enrichment Quality}

The validity of the generated semantics is examined from two complementary aspects: distributional fidelity and semantic informativeness, covering temporal, user, merchant, and textual perspectives. As shown in Figure~\ref{fig:semantic_quality}(\subref{fig:semantic_quality_time})--(\subref{fig:semantic_quality_merchant}), the generated semantics exhibit strong distributional fidelity to their corresponding references. Normal timestamps closely follow the reference hourly and weekly rhythms, while the generated age-group and provincial distributions show little deviation from the demographic priors. Merchant-category assignments also remain broadly aligned with both amount and count priors. Overall, the enrichment process preserves the major statistical characteristics.

Semantic informativeness arises from preserving fraud-relevant variation and discriminative semantic structure. In Figure~\ref{fig:semantic_quality}(\subref{fig:semantic_quality_time}), fraud timestamps exhibit distinct temporal patterns from normal transactions, with stronger activity around midday. Across the week, fraud transactions show a pronounced decline before the weekend followed by a strong weekend increase. In Figure~\ref{fig:semantic_quality}(\subref{fig:semantic_quality_text}), within-group descriptions show greater density at higher cosine similarities than cross-group descriptions, indicating semantic coherence within profiles and categories while retaining variation and cross-group discriminability.

\begin{table*}[t]
\centering
\caption{
Downstream fraud detection performance on MS-FFSD under progressive semantic enrichment. \textit{Structured} augments \textit{Original} with structured semantic attributes, while \textit{Textual} further adds entity-level textual descriptions. Positive gains over \textit{Original} are shaded, with darker colors indicating larger improvements within each metric. All results are reported in percentage (\%).}
\label{tab:downstream_fraud_detection}
\small
\setlength{\tabcolsep}{15.0pt}
\renewcommand{\arraystretch}{1.1}
\resizebox{1\textwidth}{!}{
\begin{tabular}{lccccccccc}
\toprule
& \multicolumn{3}{c}{\textbf{Original Features}}
& \multicolumn{3}{c}{\textbf{+ Structured Semantics}}
& \multicolumn{3}{c}{\textbf{+ Textual Semantics}} \\
\cmidrule(lr){2-4}
\cmidrule(lr){5-7}
\cmidrule(lr){8-10}
\textbf{Model}
& \textbf{AUC}
& \textbf{AP}
& \textbf{F1}
& \textbf{AUC}
& \textbf{AP}
& \textbf{F1}
& \textbf{AUC}
& \textbf{AP}
& \textbf{F1} \\
\midrule

GCN
& 85.30 & 96.29 & 70.04
& \gaincell{86.47}{85.30}{3.74}
& \gaincell{96.31}{96.29}{0.94}
& \gaincell{74.29}{70.04}{6.01}
& \gaincell{89.04}{85.30}{3.74}
& \gaincell{97.14}{96.29}{0.94}
& \gaincell{76.05}{70.04}{6.01} \\

GAT
& 86.92 & 96.58 & 73.47
& \gaincell{87.83}{86.92}{3.74}
& \gaincell{97.09}{96.58}{0.94}
& \gaincell{74.07}{73.47}{6.01}
& \gaincell{88.24}{86.92}{3.74}
& \gaincell{97.19}{96.58}{0.94}
& \gaincell{73.76}{73.47}{6.01} \\

GraphSAGE
& 89.43 & 97.43 & 74.86
& \gaincell{89.82}{89.43}{3.74}
& \gaincell{97.27}{97.43}{0.94}
& \gaincell{76.42}{74.86}{6.01}
& \gaincell{90.27}{89.43}{3.74}
& \gaincell{97.59}{97.43}{0.94}
& \gaincell{75.13}{74.86}{6.01} \\

CARE-GNN
& 86.03 & 96.07 & 71.02
& \gaincell{86.48}{86.03}{3.74}
& \gaincell{96.27}{96.07}{0.94}
& \gaincell{71.61}{71.02}{6.01}
& \gaincell{88.28}{86.03}{3.74}
& \gaincell{96.81}{96.07}{0.94}
& \gaincell{72.81}{71.02}{6.01} \\

PC-GNN
& 87.86 & 96.82 & 76.73
& \gaincell{87.86}{87.86}{3.74}
& \gaincell{96.80}{96.82}{0.94}
& \gaincell{76.15}{76.73}{6.01}
& \gaincell{88.94}{87.86}{3.74}
& \gaincell{96.97}{96.82}{0.94}
& \gaincell{77.89}{76.73}{6.01} \\

H2-FDetector
& 87.05 & 96.06 & 70.04
& \gaincell{87.77}{87.05}{3.74}
& \gaincell{96.93}{96.06}{0.94}
& \gaincell{71.69}{70.04}{6.01}
& \gaincell{88.41}{87.05}{3.74}
& \gaincell{97.00}{96.06}{0.94}
& \gaincell{73.12}{70.04}{6.01} \\

BWGNN
& 88.00 & 96.80 & 73.97
& \gaincell{87.34}{88.00}{3.74}
& \gaincell{96.54}{96.80}{0.94}
& \gaincell{70.95}{73.97}{6.01}
& \gaincell{89.34}{88.00}{3.74}
& \gaincell{97.14}{96.80}{0.94}
& \gaincell{76.08}{73.97}{6.01} \\

GAGA
& 85.75 & 95.62 & 75.96
& \gaincell{87.54}{85.75}{3.74}
& \gaincell{95.96}{95.62}{0.94}
& \gaincell{76.45}{75.96}{6.01}
& \gaincell{85.73}{85.75}{3.74}
& \gaincell{96.51}{95.62}{0.94}
& \gaincell{76.18}{75.96}{6.01} \\

SparseGAD
& 89.99 & 97.16 & 76.30
& \gaincell{87.75}{89.99}{3.74}
& \gaincell{96.45}{97.16}{0.94}
& \gaincell{76.12}{76.30}{6.01}
& \gaincell{90.42}{89.99}{3.74}
& \gaincell{97.47}{97.16}{0.94}
& \gaincell{77.26}{76.30}{6.01} \\

SplitGNN
& 87.42 & 96.51 & 71.81
& \gaincell{86.29}{87.42}{3.74}
& \gaincell{96.31}{96.51}{0.94}
& \gaincell{74.38}{71.81}{6.01}
& \gaincell{88.65}{87.42}{3.74}
& \gaincell{96.99}{96.51}{0.94}
& \gaincell{76.27}{71.81}{6.01} \\

ConsisGAD
& 86.57 & 96.64 & 74.27
& \gaincell{86.87}{86.57}{3.74}
& \gaincell{96.57}{96.64}{0.94}
& \gaincell{74.59}{74.27}{6.01}
& \gaincell{88.34}{86.57}{3.74}
& \gaincell{96.95}{96.64}{0.94}
& \gaincell{71.97}{74.27}{6.01} \\

PMP
& 89.40 & 97.23 & 74.77
& \gaincell{89.38}{89.40}{3.74}
& \gaincell{97.32}{97.23}{0.94}
& \gaincell{73.93}{74.77}{6.01}
& \gaincell{89.75}{89.40}{3.74}
& \gaincell{97.44}{97.23}{0.94}
& \gaincell{74.83}{74.77}{6.01} \\

DGA-GNN
& 88.76 & 97.19 & 79.17
& \gaincell{90.86}{88.76}{3.74}
& \gaincell{97.56}{97.19}{0.94}
& \gaincell{79.06}{79.17}{6.01}
& \gaincell{91.09}{88.76}{3.74}
& \gaincell{97.62}{97.19}{0.94}
& \gaincell{80.37}{79.17}{6.01} \\

SpaceGNN
& 88.77 & 97.51 & 77.60
& \gaincell{88.59}{88.77}{3.74}
& \gaincell{97.00}{97.51}{0.94}
& \gaincell{74.82}{77.60}{6.01}
& \gaincell{90.33}{88.77}{3.74}
& \gaincell{96.33}{97.51}{0.94}
& \gaincell{75.38}{77.60}{6.01} \\

CGADM
& 90.09 & 97.46 & 80.67
& \gaincell{89.76}{90.09}{3.74}
& \gaincell{97.24}{97.46}{0.94}
& \gaincell{78.34}{80.67}{6.01}
& \gaincell{90.37}{90.09}{3.74}
& \gaincell{97.33}{97.46}{0.94}
& \gaincell{80.71}{80.67}{6.01} \\

GAAP
& 89.73 & 97.35 & 76.09
& \gaincell{87.60}{89.73}{3.74}
& \gaincell{96.35}{97.35}{0.94}
& \gaincell{75.16}{76.09}{6.01}
& \gaincell{89.57}{89.73}{3.74}
& \gaincell{97.18}{97.35}{0.94}
& \gaincell{77.58}{76.09}{6.01} \\

\midrule

\textbf{Average}
& 87.94 & 96.80 & 74.80
& \gaincell{88.01}{87.94}{3.74}
& \gaincell{96.75}{96.80}{0.94}
& \gaincell{74.88}{74.80}{6.01}
& \gaincell{89.17}{87.94}{3.74}
& \gaincell{97.10}{96.80}{0.94}
& \gaincell{75.96}{74.80}{6.01} \\

\bottomrule
\end{tabular}
}
\end{table*}

\subsection{Downstream Fraud Detection with Graph Models}
\label{sec:graph_downstream}

From the predictive perspective, we examine how progressive semantic enrichment affects downstream graph-based fraud detection on MS-FFSD. Users and merchants are represented as nodes in an entity graph, with fraud detection performed on user nodes. We consider 16 representative graph-based detectors: GCN~\cite{ref13}, GAT~\cite{ref14}, GraphSAGE~\cite{ref15}, CARE-GNN~\cite{ref16}, PC-GNN~\cite{ref17}, H2-FDetector~\cite{ref18}, BWGNN~\cite{ref19}, GAGA~\cite{ref20}, SparseGAD~\cite{ref21}, SplitGNN~\cite{ref22}, ConsisGAD~\cite{ref23}, PMP~\cite{ref24}, DGA-GNN~\cite{ref25}, SpaceGNN~\cite{ref26}, CGADM~\cite{ref27}, and GAAP~\cite{ref28}. Each model is evaluated under three progressive settings: \textit{Original Features}, \textit{+ Structured Semantics}, and \textit{+ Textual Semantics}. The graph structure and experimental configuration remain fixed, with only the available semantics progressively enriched. Structured and textual semantics are encoded with Qwen3-Embedding-8B~\citep{ref30}, while numerical features are projected separately. Detailed configurations are provided in Appendix~\ref{app:graph_details}, including entity-graph construction and relation design, semantic template construction, feature encoding, and experimental settings.

Table~\ref{tab:downstream_fraud_detection} shows that the enriched semantic information in MS-FFSD improves downstream fraud detection across diverse graph architectures, with darker shading indicating larger gains over \textit{Original}. First, structured semantics produce mixed effects, with some detectors benefiting while others show marginal changes or occasional degradation, indicating architecture-dependent integration of multiple semantic signals. Second, textual semantics yield more consistent improvements and often recover performance drops introduced under structured semantics. This suggests that textual semantics enrich the structured representation with additional behavioral and business context aggregated for entity-level description generation. Third, the gains are more evident in AUC and F1-macro than in AP, suggesting that semantic enrichment primarily improves the separation of difficult or borderline samples and classification decisions, while producing smaller changes to the positive-sample ranking.

\subsection{General Applicability of the Semantic Enrichment Framework}

\begin{table}[t]
\centering
\caption{
Downstream fraud detection performance on private real-world financial datasets under progressive semantic enrichment. \textit{Structured} augments \textit{Original} with structured semantics, while \textit{Textual} further adds entity textual descriptions. Positive gains over \textit{Original} are shaded, with darker colors indicating larger improvements within each metric. Results are reported in percentage (\%).
}
\label{tab:realworld_downstream}
\scriptsize
\setlength{\tabcolsep}{2pt}
\renewcommand{\arraystretch}{1.1}

\begin{tabular}{lccccccccc|ccccccccc}
\toprule

& \multicolumn{9}{c|}{\textbf{Private-1}}
& \multicolumn{9}{c}{\textbf{Private-2}} \\

\cmidrule(lr){2-10}
\cmidrule(lr){11-19}

& \multicolumn{3}{c}{\textbf{Original}}
& \multicolumn{3}{c}{\textbf{+ Structured}}
& \multicolumn{3}{c|}{\textbf{+ Textual}}

& \multicolumn{3}{c}{\textbf{Original}}
& \multicolumn{3}{c}{\textbf{+ Structured}}
& \multicolumn{3}{c}{\textbf{+ Textual}} \\

\cmidrule(lr){2-4}
\cmidrule(lr){5-7}
\cmidrule(lr){8-10}
\cmidrule(lr){11-13}
\cmidrule(lr){14-16}
\cmidrule(lr){17-19}

\textbf{Model}
& \textbf{AUC} & \textbf{AP} & \textbf{F1}
& \textbf{AUC} & \textbf{AP} & \textbf{F1}
& \textbf{AUC} & \textbf{AP} & \textbf{F1}
& \textbf{AUC} & \textbf{AP} & \textbf{F1}
& \textbf{AUC} & \textbf{AP} & \textbf{F1}
& \textbf{AUC} & \textbf{AP} & \textbf{F1} \\

\midrule

GCN
& 90.36 & 98.84 & 70.09
& \gaincell{93.73}{90.36}{3.89}
& \gaincell{99.28}{98.84}{2.32}
& \gaincell{76.82}{70.09}{15.07}
& \gaincell{94.17}{90.36}{3.89}
& \gaincell{99.47}{98.84}{2.32}
& \gaincell{78.39}{70.09}{15.07}
& 96.43 & 98.98 & 87.21
& \gaincell{96.41}{96.43}{1.72}
& \gaincell{98.97}{98.98}{2.24}
& \gaincell{88.69}{87.21}{4.62}
& \gaincell{97.90}{96.43}{1.72}
& \gaincell{99.41}{98.98}{2.24}
& \gaincell{89.88}{87.21}{4.62} \\

GAT
& 93.43 & 99.26 & 75.34
& \gaincell{94.22}{93.43}{3.89}
& \gaincell{99.40}{99.26}{2.32}
& \gaincell{77.32}{75.34}{15.07}
& \gaincell{94.68}{93.43}{3.89}
& \gaincell{99.41}{99.26}{2.32}
& \gaincell{77.99}{75.34}{15.07}
& 96.85 & 99.04 & 89.58
& \gaincell{97.00}{96.85}{1.72}
& \gaincell{99.08}{99.04}{2.24}
& \gaincell{90.21}{89.58}{4.62}
& \gaincell{97.36}{96.85}{1.72}
& \gaincell{99.20}{99.04}{2.24}
& \gaincell{91.24}{89.58}{4.62} \\

GraphSAGE
& 94.66 & 99.42 & 76.08
& \gaincell{94.67}{94.66}{3.89}
& \gaincell{99.40}{99.42}{2.32}
& \gaincell{77.61}{76.08}{15.07}
& \gaincell{95.04}{94.66}{3.89}
& \gaincell{99.45}{99.42}{2.32}
& \gaincell{78.16}{76.08}{15.07}
& 98.67 & 99.63 & 92.95
& \gaincell{98.51}{98.67}{1.72}
& \gaincell{99.59}{99.63}{2.24}
& \gaincell{93.16}{92.95}{4.62}
& \gaincell{98.76}{98.67}{1.72}
& \gaincell{99.64}{99.63}{2.24}
& \gaincell{93.18}{92.95}{4.62} \\

CARE-GNN
& 90.10 & 98.31 & 64.96
& \gaincell{92.05}{90.10}{3.89}
& \gaincell{99.09}{98.31}{2.32}
& \gaincell{71.71}{64.96}{15.07}
& \gaincell{93.99}{90.10}{3.89}
& \gaincell{99.34}{98.31}{2.32}
& \gaincell{77.83}{64.96}{15.07}
& 97.75 & 99.38 & 91.50
& \gaincell{98.02}{97.75}{1.72}
& \gaincell{99.43}{99.38}{2.24}
& \gaincell{90.75}{91.50}{4.62}
& \gaincell{98.06}{97.75}{1.72}
& \gaincell{99.44}{99.38}{2.24}
& \gaincell{92.54}{91.50}{4.62} \\

PC-GNN
& 90.66 & 97.82 & 61.85
& \gaincell{91.05}{90.66}{3.89}
& \gaincell{98.83}{97.82}{2.32}
& \gaincell{75.74}{61.85}{15.07}
& \gaincell{93.54}{90.66}{3.89}
& \gaincell{99.26}{97.82}{2.32}
& \gaincell{75.72}{61.85}{15.07}
& 97.81 & 99.40 & 90.34
& \gaincell{98.51}{97.81}{1.72}
& \gaincell{99.59}{99.40}{2.24}
& \gaincell{90.04}{90.34}{4.62}
& \gaincell{98.25}{97.81}{1.72}
& \gaincell{99.52}{99.40}{2.24}
& \gaincell{92.13}{90.34}{4.62} \\

H2-FDetector
& 90.78 & 97.07 & 71.69
& \gaincell{93.82}{90.78}{3.89}
& \gaincell{99.29}{97.07}{2.32}
& \gaincell{73.78}{71.69}{15.07}
& \gaincell{93.66}{90.78}{3.89}
& \gaincell{99.39}{97.07}{2.32}
& \gaincell{73.67}{71.69}{15.07}
& 97.56 & 99.31 & 86.76
& \gaincell{96.71}{97.56}{1.72}
& \gaincell{99.03}{99.31}{2.24}
& \gaincell{86.75}{86.76}{4.62}
& \gaincell{97.02}{97.56}{1.72}
& \gaincell{99.43}{99.31}{2.24}
& \gaincell{88.49}{86.76}{4.62} \\

BWGNN
& 93.47 & 99.23 & 75.63
& \gaincell{92.65}{93.47}{3.89}
& \gaincell{99.14}{99.23}{2.32}
& \gaincell{71.12}{75.63}{15.07}
& \gaincell{94.10}{93.47}{3.89}
& \gaincell{99.36}{99.23}{2.32}
& \gaincell{76.02}{75.63}{15.07}
& 97.75 & 99.36 & 91.13
& \gaincell{97.75}{97.75}{1.72}
& \gaincell{99.38}{99.36}{2.24}
& \gaincell{90.96}{91.13}{4.62}
& \gaincell{98.31}{97.75}{1.72}
& \gaincell{99.52}{99.36}{2.24}
& \gaincell{92.38}{91.13}{4.62} \\

GAGA
& 92.82 & 99.18 & 78.32
& \gaincell{92.55}{92.82}{3.89}
& \gaincell{98.95}{99.18}{2.32}
& \gaincell{67.14}{78.32}{15.07}
& \gaincell{92.93}{92.82}{3.89}
& \gaincell{99.24}{99.18}{2.32}
& \gaincell{79.35}{78.32}{15.07}
& 97.99 & 99.46 & 91.37
& \gaincell{98.11}{97.99}{1.72}
& \gaincell{99.48}{99.46}{2.24}
& \gaincell{92.84}{91.37}{4.62}
& \gaincell{98.18}{97.99}{1.72}
& \gaincell{99.50}{99.46}{2.24}
& \gaincell{91.96}{91.37}{4.62} \\

SparseGAD
& 93.98 & 99.34 & 73.83
& \gaincell{94.40}{93.98}{3.89}
& \gaincell{99.37}{99.34}{2.32}
& \gaincell{73.54}{73.83}{15.07}
& \gaincell{94.87}{93.98}{3.89}
& \gaincell{99.44}{99.34}{2.32}
& \gaincell{78.88}{73.83}{15.07}
& 97.95 & 99.40 & 89.71
& \gaincell{97.87}{97.95}{1.72}
& \gaincell{99.42}{99.40}{2.24}
& \gaincell{90.96}{89.71}{4.62}
& \gaincell{98.39}{97.95}{1.72}
& \gaincell{99.56}{99.40}{2.24}
& \gaincell{91.30}{89.71}{4.62} \\

SplitGNN
& 93.22 & 99.33 & 77.91
& \gaincell{94.11}{93.22}{3.89}
& \gaincell{99.23}{99.33}{2.32}
& \gaincell{76.16}{77.91}{15.07}
& \gaincell{93.91}{93.22}{3.89}
& \gaincell{99.38}{99.33}{2.32}
& \gaincell{76.63}{77.91}{15.07}
& 97.48 & 99.10 & 91.28
& \gaincell{97.43}{97.48}{1.72}
& \gaincell{99.29}{99.10}{2.24}
& \gaincell{91.48}{91.28}{4.62}
& \gaincell{97.82}{97.48}{1.72}
& \gaincell{99.41}{99.10}{2.24}
& \gaincell{91.19}{91.28}{4.62} \\

ConsisGAD
& 92.40 & 99.16 & 65.14
& \gaincell{92.58}{92.40}{3.89}
& \gaincell{99.17}{99.16}{2.32}
& \gaincell{66.28}{65.14}{15.07}
& \gaincell{93.55}{92.40}{3.89}
& \gaincell{99.22}{99.16}{2.32}
& \gaincell{80.21}{65.14}{15.07}
& 95.36 & 98.15 & 85.29
& \gaincell{95.58}{95.36}{1.72}
& \gaincell{98.17}{98.15}{2.24}
& \gaincell{86.92}{85.29}{4.62}
& \gaincell{96.56}{95.36}{1.72}
& \gaincell{99.27}{98.15}{2.24}
& \gaincell{87.37}{85.29}{4.62} \\

PMP
& 95.60 & 99.51 & 80.03
& \gaincell{95.33}{95.60}{3.89}
& \gaincell{99.45}{99.51}{2.32}
& \gaincell{78.65}{80.03}{15.07}
& \gaincell{95.72}{95.60}{3.89}
& \gaincell{99.66}{99.51}{2.32}
& \gaincell{78.16}{80.03}{15.07}
& 98.32 & 99.55 & 88.19
& \gaincell{98.26}{98.32}{1.72}
& \gaincell{99.52}{99.55}{2.24}
& \gaincell{88.48}{88.19}{4.62}
& \gaincell{98.36}{98.32}{1.72}
& \gaincell{99.57}{99.55}{2.24}
& \gaincell{92.81}{88.19}{4.62} \\

DGA-GNN
& 93.92 & 99.32 & 78.22
& \gaincell{94.38}{93.92}{3.89}
& \gaincell{99.38}{99.32}{2.32}
& \gaincell{78.71}{78.22}{15.07}
& \gaincell{95.98}{93.92}{3.89}
& \gaincell{99.21}{99.32}{2.32}
& \gaincell{79.65}{78.22}{15.07}
& 97.79 & 99.39 & 91.28
& \gaincell{97.83}{97.79}{1.72}
& \gaincell{99.43}{99.39}{2.24}
& \gaincell{91.86}{91.28}{4.62}
& \gaincell{98.40}{97.79}{1.72}
& \gaincell{99.54}{99.39}{2.24}
& \gaincell{92.68}{91.28}{4.62} \\

SpaceGNN
& 94.78 & 99.42 & 74.82
& \gaincell{94.43}{94.78}{3.89}
& \gaincell{99.35}{99.42}{2.32}
& \gaincell{75.64}{74.82}{15.07}
& \gaincell{95.00}{94.78}{3.89}
& \gaincell{99.45}{99.42}{2.32}
& \gaincell{74.54}{74.82}{15.07}
& 98.51 & 99.59 & 91.19
& \gaincell{98.45}{98.51}{1.72}
& \gaincell{99.60}{99.59}{2.24}
& \gaincell{93.64}{91.19}{4.62}
& \gaincell{98.71}{98.51}{1.72}
& \gaincell{99.65}{99.59}{2.24}
& \gaincell{94.16}{91.19}{4.62} \\

CGADM
& 95.42 & 99.49 & 79.80
& \gaincell{95.31}{95.42}{3.89}
& \gaincell{99.47}{99.49}{2.32}
& \gaincell{78.87}{79.80}{15.07}
& \gaincell{95.46}{95.42}{3.89}
& \gaincell{99.52}{99.49}{2.32}
& \gaincell{79.28}{79.80}{15.07}
& 96.86 & 97.38 & 93.04
& \gaincell{98.17}{96.86}{1.72}
& \gaincell{98.75}{97.38}{2.24}
& \gaincell{93.14}{93.04}{4.62}
& \gaincell{98.58}{96.86}{1.72}
& \gaincell{99.62}{97.38}{2.24}
& \gaincell{94.25}{93.04}{4.62} \\

GAAP
& 94.74 & 99.37 & 80.40
& \gaincell{94.98}{94.74}{3.89}
& \gaincell{99.38}{99.37}{2.32}
& \gaincell{81.51}{80.40}{15.07}
& \gaincell{95.31}{94.74}{3.89}
& \gaincell{99.10}{99.37}{2.32}
& \gaincell{82.68}{80.40}{15.07}
& 97.83 & 99.40 & 92.38
& \gaincell{98.19}{97.83}{1.72}
& \gaincell{99.49}{99.40}{2.24}
& \gaincell{93.93}{92.38}{4.62}
& \gaincell{98.77}{97.83}{1.72}
& \gaincell{99.65}{99.40}{2.24}
& \gaincell{94.16}{92.38}{4.62} \\

\midrule

\textbf{Average}
& 93.15 & 99.00 & 74.01
& \gaincell{93.77}{93.15}{3.89}
& \gaincell{99.26}{99.00}{2.32}
& \gaincell{75.04}{74.01}{15.07}
& \gaincell{94.49}{93.15}{3.89}
& \gaincell{99.37}{99.00}{2.32}
& \gaincell{77.95}{74.01}{15.07}
& 97.56 & 99.16 & 90.20
& \gaincell{97.68}{97.56}{1.72}
& \gaincell{99.26}{99.16}{2.24}
& \gaincell{90.86}{90.20}{4.62}
& \gaincell{98.09}{97.56}{1.72}
& \gaincell{99.50}{99.16}{2.24}
& \gaincell{91.86}{90.20}{4.62} \\

\bottomrule
\end{tabular}

\end{table}

To establish the broader applicability of the proposed semantic enrichment framework, we further examine whether its predictive benefits can be consistently reproduced across datasets with different characteristics. We apply the enrichment framework to two private real-world financial transaction datasets. Further dataset statistics and descriptions are provided in Appendix~\ref{app:private_data}. The generated structured and textual semantics are evaluated following the progressive protocol in Section~\ref{sec:graph_downstream}.

Table~\ref{tab:realworld_downstream} shows that the semantics generated by the framework improve downstream fraud detection on both datasets. Structured semantics exhibit model-dependent effects, whereas incorporating textual semantics leads to broader and more consistent improvements across detectors. Notably, the gains persist even when the original features already provide strong predictive performance, suggesting that the generated semantics add predictive information beyond the existing feature space. The recurring improvement pattern across independently enriched datasets provides empirical evidence for the effectiveness and broader applicability of the proposed framework.

\subsection{LLM-based Semantic Reasoning}

From the reasoning perspective, we conduct zero-shot semantic reasoning with DeepSeek-V3~\citep{ref52} under progressive enrichment settings~\citep{ref40,ref41}. We use a qualitative case-based analysis to examine whether richer semantics support contextual reasoning over transaction behavior. Original features, structured and textual semantics are introduced in sequence. At each level, the LLM and graph models receive the same underlying information but in different representations: graph models use projected numerical features and text embeddings, whereas the LLM directly reads the corresponding numerical values and textual semantics. Setup, prompts, and case selection criteria are detailed in Appendix~\ref{app:llm_reasoning}.

Figure~\ref{fig:llm_reasoning_cases} presents illustrative examples illustrating how progressive semantic enrichment changes the LLM's interpretation of transaction behavior. With original features, the reasoning is primarily driven by numerical patterns and observed transaction variations. Structured semantics ground these observations in explicit user and merchant attributes, while textual semantics further integrate them into contextual behavioral interpretations. In Example A, this richer context helps reconcile numerical variations with plausible behavior. In Example B, part of the variation becomes contextually explainable, while a residual anomaly remains inconsistent with the entity profile. Together, these cases show that textual semantics support contextual interpretation by integrating transaction patterns, entity attributes, and behavioral context, while preserving sensitivity to unexplained abnormalities.

\begin{figure}[t]
  \centering
  \includegraphics[width=1\linewidth]{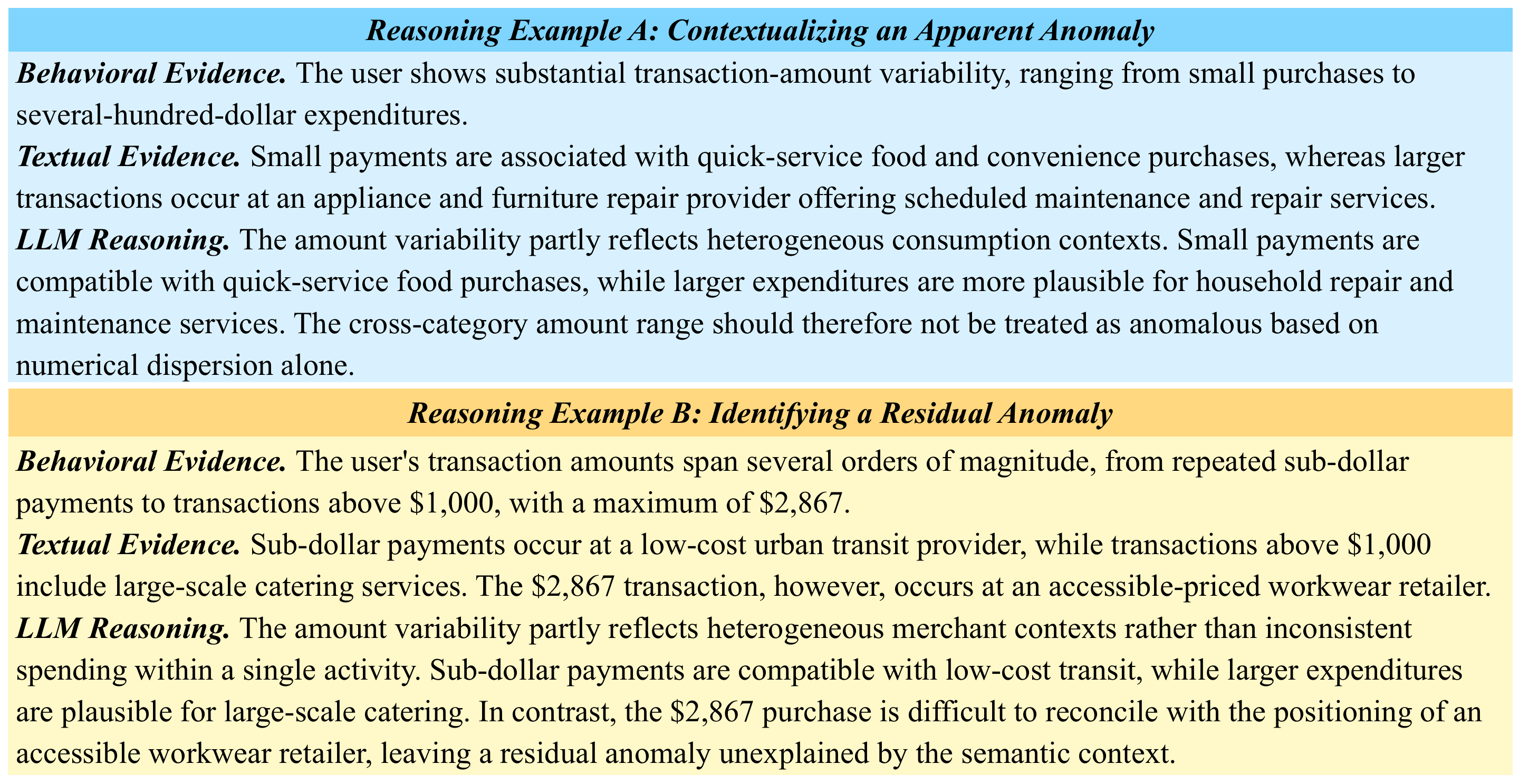}
  \caption{Illustrative examples of LLM-based semantic reasoning. Example A shows how rich-text semantics support context-aware interpretation of numerical variation, while Example B shows how semantic reasoning distinguishes contextually plausible variation from a residual anomaly.}
  \label{fig:llm_reasoning_cases}
\end{figure}

\section{Conclusion}

In this work, we proposed a multi-agent semantic enrichment framework for financial fraud detection and newly contributed a valuable multimodal fraud detection dataset MS-FFSD that enriches with structured and textual semantics grounded in observed transaction behavior. Our evaluation demonstrates the statistical fidelity and generalizability of the enrichment framework, while showing the value of richer semantics for financial fraud modeling and contextual reasoning. These results highlight the potential of behavior-grounded semantic enrichment for advancing multimodal financial fraud research. Furthermore, our work facilitates systematic probing of the capabilities and boundaries of LLMs and multi-agent systems in realistic and complex financial settings. Crucially, it establishes a bridge between academic advances and anti-fraud practice, positioning emerging modeling and reasoning capabilities for operational financial fraud prevention.

\subsection*{AI use statement}

In this work, we used generative AI tools as part of the proposed semantic enrichment framework to generate textual semantics for the released dataset. Generative AI was also used during research development to assist with methodological discussion, mathematical formulation, experimental analysis, and interpretation of results. In addition, we used generative AI for literature analysis, brainstorming, drafting and editing parts of the manuscript, improving readability, and refining the presentation and structure of the paper. All AI-assisted methodological decisions, generated data, mathematical formulations, experimental interpretations, and manuscript content were reviewed and verified by the authors. We take responsibility for the final content of this work, including text, claims or artifacts produced with the aid of generative AI.

\subsection*{Ethics Statement}

This work focuses on semantic enrichment for financial fraud research and does not involve direct interaction with human subjects. The released dataset does not contain personally identifiable information, and the enriched demographic, geographic, merchant, and textual semantics are synthetically constructed rather than observations of real individuals. These generated attributes should therefore not be interpreted as factual profiles of specific persons. The released dataset and framework are intended for research purposes.

\subsection*{Reproducibility Statement}

We provide detailed documentation and open-source resources to support reproducibility of both the semantic enrichment framework and the resulting dataset. An anonymous repository containing the released dataset and the complete implementation of the semantic enrichment framework is linked in the abstract. The main paper describes the overall framework, dataset organization, and evaluation protocols. Appendix~\ref{app:framework_details} provides the mathematical formulation and agent-level objectives, Appendix~\ref{app:external_knowledge} documents the external knowledge sources and their roles, and Appendix~\ref{app:dataset_details} provides detailed definitions and statistics of the enriched semantic attributes. Appendix~\ref{app:private_data} describes the two private transaction datasets used for framework evaluation. The prompts and deterministic templates used for textual generation are provided in Appendix~\ref{app:text_generation_prompts}. Detailed settings for graph-based fraud detection and LLM reasoning are reported in Appendices~\ref{app:graph_details} and~\ref{app:llm_reasoning}, respectively.



\bibliography{refs}

@article{ref2,
title = {Credit card fraud detection in the era of disruptive technologies: A systematic review},
journal = {Journal of King Saud University - Computer and Information Sciences},
volume = {35},
number = {1},
pages = {145-174},
year = {2023},
issn = {1319-1578},
author = {Asma Cherif and Arwa Badhib and Heyfa Ammar and Suhair Alshehri and Manal Kalkatawi and Abdessamad Imine}
}

@article{ref51,
  title={Graph neural networks for financial fraud detection: a review},
  author={Cheng, Dawei and Zou, Yao and Xiang, Sheng and Jiang, Changjun},
  journal={Frontiers of Computer Science},
  volume={19},
  number={9},
  pages={199609},
  year={2025},
  publisher={Springer}
}

@misc{ref3,
      title={Fraud Dataset Benchmark and Applications}, 
      author={Prince Grover and Julia Xu and Justin Tittelfitz and Anqi Cheng and Zheng Li and Jakub Zablocki and Jianbo Liu and Hao Zhou},
      year={2023},
      eprint={2208.14417},
      archivePrefix={arXiv},
      primaryClass={cs.LG}
}

@ARTICLE{ref33,
  author={Ma, Xiaoxiao and Wu, Jia and Xue, Shan and Yang, Jian and Zhou, Chuan and Sheng, Quan Z. and Xiong, Hui and Akoglu, Leman},
  journal={IEEE Transactions on Knowledge and Data Engineering}, 
  title={A Comprehensive Survey on Graph Anomaly Detection With Deep Learning}, 
  year={2023},
  volume={35},
  number={12},
  pages={12012-12038}}

@inproceedings{ref4,
author = {Altman, Erik},
title = {Synthesizing credit card transactions},
year = {2022},
isbn = {9781450391481},
publisher = {Association for Computing Machinery},
address = {New York, NY, USA},
booktitle = {Proceedings of the Second ACM International Conference on AI in Finance},
articleno = {13},
numpages = {9},
location = {Virtual Event},
series = {ICAIF '21}
}

@INPROCEEDINGS{ref9,
  author={Padhi, Inkit and Schiff, Yair and Melnyk, Igor and Rigotti, Mattia and Mroueh, Youssef and Dognin, Pierre and Ross, Jerret and Nair, Ravi and Altman, Erik},
  booktitle={ICASSP 2021 - 2021 IEEE International Conference on Acoustics, Speech and Signal Processing (ICASSP)}, 
  title={Tabular Transformers for Modeling Multivariate Time Series}, 
  year={2021},
  volume={},
  number={},
  pages={3565-3569}}

@misc{ref10,
  author       = {Erik Altman},
  title        = {{Credit Card Transactions}},
  year         = {2019},
  howpublished = {Kaggle Dataset},
  url          = {https://www.kaggle.com/datasets/ealtman2019/credit-card-transactions}
}

@misc{ref5,
  author       = {{Vesta Corporation}},
  title        = {{IEEE-CIS Fraud Detection}},
  year         = {2019},
  howpublished = {Kaggle Competition},
  url          = {https://www.kaggle.com/c/ieee-fraud-detection}
}

@misc{ref6,
  author       = {Binh Vu},
  title        = {{Fraud Ecommerce}},
  year         = {2018},
  howpublished = {Kaggle Dataset},
  url          = {https://www.kaggle.com/datasets/vbinh002/fraud-ecommerce}
}

@INPROCEEDINGS{ref7,
  author={Pozzolo, Andrea Dal and Caelen, Olivier and Johnson, Reid A. and Bontempi, Gianluca},
  booktitle={2015 IEEE Symposium Series on Computational Intelligence}, 
  title={Calibrating Probability with Undersampling for Unbalanced Classification}, 
  year={2015},
  volume={},
  number={},
  pages={159-166}}

@misc{ref8,
  author       = {Kartik Shenoy},
  title        = {{Credit Card Transactions Fraud Detection Dataset}},
  year         = {2020},
  howpublished = {Kaggle Dataset},
  url          = {https://www.kaggle.com/datasets/kartik2112/fraud-detection}
}

@INPROCEEDINGS{ref32,
  author={Oztas, Berkan and Cetinkaya, Deniz and Adedoyin, Festus and Budka, Marcin and Dogan, Huseyin and Aksu, Gokhan},
  booktitle={2023 IEEE International Conference on e-Business Engineering (ICEBE)}, 
  title={Enhancing Anti-Money Laundering: Development of a Synthetic Transaction Monitoring Dataset}, 
  year={2023},
  volume={},
  number={},
  pages={47-54}}

@inproceedings{ref37,
 author = {Altman, Erik and Blanu\v{s}a, Jovan and von Niederh\"{a}usern, Luc and Egressy, Beni and Anghel, Andreea and Atasu, Kubilay},
 booktitle = {Advances in Neural Information Processing Systems},
 editor = {A. Oh and T. Naumann and A. Globerson and K. Saenko and M. Hardt and S. Levine},
 pages = {29851--29874},
 publisher = {Curran Associates, Inc.},
 title = {Realistic Synthetic Financial Transactions for Anti-Money Laundering Models},
 volume = {36},
 year = {2023}
}

@inproceedings{ref31,
author = {Elmougy, Youssef and Liu, Ling},
title = {Demystifying Fraudulent Transactions and Illicit Nodes in the Bitcoin Network for Financial Forensics},
year = {2023},
isbn = {9798400701030},
publisher = {Association for Computing Machinery},
address = {New York, NY, USA},
booktitle = {Proceedings of the 29th ACM SIGKDD Conference on Knowledge Discovery and Data Mining},
pages = {3979–3990},
numpages = {12},
location = {Long Beach, CA, USA},
series = {KDD '23}
}

@inproceedings{ref11,
  title={Semi-supervised credit card fraud detection via attribute-driven graph representation},
  author={Xiang, Sheng and Zhu, Mingzhi and Cheng, Dawei and Li, Enxia and Zhao, Ruihui and Ouyang, Yi and Chen, Ling and Zheng, Yefeng},
  booktitle={Proceedings of the AAAI Conference on Artificial Intelligence},
  volume={37},
  pages={14557--14565},
  year={2023}
}

@inproceedings{ref12,
  title={Efficient dynamic attributed graph generation},
  author={Li, Fan and Wang, Xiaoyang and Cheng, Dawei and Chen, Cong and Zhang, Ying and Lin, Xuemin},
  booktitle={2025 IEEE 41st International Conference on Data Engineering (ICDE)},
  pages={1415--1428},
  year={2025},
  organization={IEEE}
}

@inproceedings{ref13,
title={Semi-Supervised Classification with Graph Convolutional Networks},
author={Thomas N. Kipf and Max Welling},
booktitle={International Conference on Learning Representations},
year={2017}
}

@inproceedings{ref14,
title={Graph Attention Networks},
author={Petar Veličković and Guillem Cucurull and Arantxa Casanova and Adriana Romero and Pietro Liò and Yoshua Bengio},
booktitle={International Conference on Learning Representations},
year={2018},
}

@inproceedings{ref15,
author = {Hamilton, William L. and Ying, Rex and Leskovec, Jure},
title = {Inductive representation learning on large graphs},
year = {2017},
isbn = {9781510860964},
publisher = {Curran Associates Inc.},
address = {Red Hook, NY, USA},
booktitle = {Proceedings of the 31st International Conference on Neural Information Processing Systems},
pages = {1025–1035},
numpages = {11},
location = {Long Beach, California, USA},
series = {NIPS'17}
}

@inproceedings{ref16,
author = {Dou, Yingtong and Liu, Zhiwei and Sun, Li and Deng, Yutong and Peng, Hao and Yu, Philip S.},
title = {Enhancing Graph Neural Network-based Fraud Detectors against Camouflaged Fraudsters},
year = {2020},
isbn = {9781450368599},
publisher = {Association for Computing Machinery},
address = {New York, NY, USA},
booktitle = {Proceedings of the 29th ACM International Conference on Information \& Knowledge Management},
pages = {315–324},
numpages = {10},
location = {Virtual Event, Ireland},
series = {CIKM '20}
}

@inproceedings{ref17,
author = {Liu, Yang and Ao, Xiang and Qin, Zidi and Chi, Jianfeng and Feng, Jinghua and Yang, Hao and He, Qing},
title = {Pick and Choose: A GNN-based Imbalanced Learning Approach for Fraud Detection},
year = {2021},
isbn = {9781450383127},
publisher = {Association for Computing Machinery},
address = {New York, NY, USA},
booktitle = {Proceedings of the Web Conference 2021},
pages = {3168–3177},
numpages = {10},
location = {Ljubljana, Slovenia},
series = {WWW '21}
}

@inproceedings{ref18,
author = {Shi, Fengzhao and Cao, Yanan and Shang, Yanmin and Zhou, Yuchen and Zhou, Chuan and Wu, Jia},
title = {H2-FDetector: A GNN-based Fraud Detector with Homophilic and Heterophilic Connections},
year = {2022},
isbn = {9781450390965},
publisher = {Association for Computing Machinery},
address = {New York, NY, USA},
booktitle = {Proceedings of the ACM Web Conference 2022},
pages = {1486–1494},
numpages = {9},
location = {Virtual Event, Lyon, France},
series = {WWW '22}
}

@InProceedings{ref19,
  title={Rethinking Graph Neural Networks for Anomaly Detection},
  author={Tang, Jianheng and Li, Jiajin and Gao, Ziqi and Li, Jia},
  booktitle={International Conference on Machine Learning},
  year={2022},
}

@inproceedings{ref20,
  title={Label information enhanced fraud detection against low homophily in graphs},
  author={Wang, Yuchen and Zhang, Jinghui and Huang, Zhengjie and Li, Weibin and Feng, Shikun and Ma, Ziheng and Sun, Yu and Yu, Dianhai and Dong, Fang and Jin, Jiahui and others},
  booktitle={Proceedings of the ACM Web Conference 2023},
  pages={406--416},
  year={2023}
}

@inproceedings{ref21,
author = {Gong, Zheng and Wang, Guifeng and Sun, Ying and Liu, Qi and Ning, Yuting and Xiong, Hui and Peng, Jingyu},
title = {Beyond homophily: robust graph anomaly detection via neural sparsification},
year = {2023},
isbn = {978-1-956792-03-4},
booktitle = {Proceedings of the Thirty-Second International Joint Conference on Artificial Intelligence},
articleno = {234},
numpages = {10},
location = {Macao, P.R.China},
series = {IJCAI '23}
}

@inproceedings{ref22,
  author = {Wu, Bin and Yao, Xinyu and Zhang, Boyan and Chao, Kuo-Ming and Li, Yinsheng},
  title = {SplitGNN: Spectral Graph Neural Network for Fraud Detection against Heterophily},
  year = {2023},
  booktitle = {Proceedings of the 32nd ACM International Conference on Information and Knowledge Management},
  series = {CIKM '23}
}

@inproceedings{ref23,
title={Consistency Training with Learnable Data Augmentation for Graph Anomaly Detection with Limited Supervision},
author={Nan Chen and Zemin Liu and Bryan Hooi and Bingsheng He and Rizal Fathony and Jun Hu and Jia Chen},
booktitle={The Twelfth International Conference on Learning Representations},
year={2024}
}

@inproceedings{ref24,
title={Partitioning Message Passing for Graph Fraud Detection},
author={Wei Zhuo and Zemin Liu and Bryan Hooi and Bingsheng He and Guang Tan and Rizal Fathony and Jia Chen},
booktitle={The Twelfth International Conference on Learning Representations},
year={2024}
}

@inproceedings{ref25,
  title={DGA-GNN: Dynamic Grouping Aggregation GNN for Fraud Detection},
  author={Duan, Mingjiang and Zheng, Tongya and Gao, Yang and Feng, Zunlei and Wang, Xinyu},
  booktitle={Proceedings of the AAAI Conference on Artificial Intelligence},
  year={2024}
}

@inproceedings{ref26,
title={Space{GNN}: Multi-Space Graph Neural Network for Node Anomaly Detection with Extremely Limited Labels},
author={Xiangyu Dong and Xingyi Zhang and Lei Chen and Mingxuan Yuan and Sibo Wang},
booktitle={The Thirteenth International Conference on Learning Representations},
year={2025}
}

@inproceedings{ref27,
title={Conditional Diffusion Anomaly Modeling on Graphs},
author={Chunyu Wei and Haozhe Lin and Yueguo Chen and Yunhai Wang},
booktitle={The Thirty-ninth Annual Conference on Neural Information Processing Systems},
year={2025}
}

@article{ref28, 
title={Global Attribute-Association Pattern Aggregation for Graph Fraud Detection}, 
volume={39},  
number={11}, 
journal={Proceedings of the AAAI Conference on Artificial Intelligence}, 
author={Duan, Mingjiang and He, Da and Zheng, Tongya and Jia, Lingxiang and Song, Mingli and Wang, Xinyu and Feng, Zunlei}, 
year={2025}, 
month={Apr.}, 
pages={11616-11624} }

@article{ref29,
    title={Qwen3 Technical Report}, 
    author={An Yang and Anfeng Li and Baosong Yang and Beichen Zhang and Binyuan Hui and Bo Zheng and Bowen Yu and Chang Gao and Chengen Huang and Chenxu Lv and Chujie Zheng and Dayiheng Liu and Fan Zhou and Fei Huang and Feng Hu and Hao Ge and Haoran Wei and Huan Lin and Jialong Tang and Jian Yang and Jianhong Tu and Jianwei Zhang and Jianxin Yang and Jiaxi Yang and Jing Zhou and Jingren Zhou and Junyang Lin and Kai Dang and Keqin Bao and Kexin Yang and Le Yu and Lianghao Deng and Mei Li and Mingfeng Xue and Mingze Li and Pei Zhang and Peng Wang and Qin Zhu and Rui Men and Ruize Gao and Shixuan Liu and Shuang Luo and Tianhao Li and Tianyi Tang and Wenbiao Yin and Xingzhang Ren and Xinyu Wang and Xinyu Zhang and Xuancheng Ren and Yang Fan and Yang Su and Yichang Zhang and Yinger Zhang and Yu Wan and Yuqiong Liu and Zekun Wang and Zeyu Cui and Zhenru Zhang and Zhipeng Zhou and Zihan Qiu},
    journal = {arXiv preprint arXiv:2505.09388},
    year={2025}
}

@article{ref30,
  title={Qwen3 Embedding: Advancing Text Embedding and Reranking Through Foundation Models},
  author={Zhang, Yanzhao and Li, Mingxin and Long, Dingkun and Zhang, Xin and Lin, Huan and Yang, Baosong and Xie, Pengjun and Yang, An and Liu, Dayiheng and Lin, Junyang and Huang, Fei and Zhou, Jingren},
  journal={arXiv preprint arXiv:2506.05176},
  year={2025}
}

@misc{ref52,
      title={DeepSeek-V3 Technical Report}, 
      author={DeepSeek-AI and Aixin Liu and Bei Feng and Bing Xue and Bingxuan Wang and Bochao Wu and Chengda Lu and Chenggang Zhao and Chengqi Deng and Chenyu Zhang and Chong Ruan and Damai Dai and Daya Guo and Dejian Yang and Deli Chen and Dongjie Ji and Erhang Li and Fangyun Lin and Fucong Dai and Fuli Luo and Guangbo Hao and Guanting Chen and Guowei Li and H. Zhang and Han Bao and Hanwei Xu and Haocheng Wang and Haowei Zhang and Honghui Ding and Huajian Xin and Huazuo Gao and Hui Li and Hui Qu and J. L. Cai and Jian Liang and Jianzhong Guo and Jiaqi Ni and Jiashi Li and Jiawei Wang and Jin Chen and Jingchang Chen and Jingyang Yuan and Junjie Qiu and Junlong Li and Junxiao Song and Kai Dong and Kai Hu and Kaige Gao and Kang Guan and Kexin Huang and Kuai Yu and Lean Wang and Lecong Zhang and Lei Xu and Leyi Xia and Liang Zhao and Litong Wang and Liyue Zhang and Meng Li and Miaojun Wang and Mingchuan Zhang and Minghua Zhang and Minghui Tang and Mingming Li and Ning Tian and Panpan Huang and Peiyi Wang and Peng Zhang and Qiancheng Wang and Qihao Zhu and Qinyu Chen and Qiushi Du and R. J. Chen and R. L. Jin and Ruiqi Ge and Ruisong Zhang and Ruizhe Pan and Runji Wang and Runxin Xu and Ruoyu Zhang and Ruyi Chen and S. S. Li and Shanghao Lu and Shangyan Zhou and Shanhuang Chen and Shaoqing Wu and Shengfeng Ye and Shengfeng Ye and Shirong Ma and Shiyu Wang and Shuang Zhou and Shuiping Yu and Shunfeng Zhou and Shuting Pan and T. Wang and Tao Yun and Tian Pei and Tianyu Sun and W. L. Xiao and Wangding Zeng and Wanjia Zhao and Wei An and Wen Liu and Wenfeng Liang and Wenjun Gao and Wenqin Yu and Wentao Zhang and X. Q. Li and Xiangyue Jin and Xianzu Wang and Xiao Bi and Xiaodong Liu and Xiaohan Wang and Xiaojin Shen and Xiaokang Chen and Xiaokang Zhang and Xiaosha Chen and Xiaotao Nie and Xiaowen Sun and Xiaoxiang Wang and Xin Cheng and Xin Liu and Xin Xie and Xingchao Liu and Xingkai Yu and Xinnan Song and Xinxia Shan and Xinyi Zhou and Xinyu Yang and Xinyuan Li and Xuecheng Su and Xuheng Lin and Y. K. Li and Y. Q. Wang and Y. X. Wei and Y. X. Zhu and Yang Zhang and Yanhong Xu and Yanhong Xu and Yanping Huang and Yao Li and Yao Zhao and Yaofeng Sun and Yaohui Li and Yaohui Wang and Yi Yu and Yi Zheng and Yichao Zhang and Yifan Shi and Yiliang Xiong and Ying He and Ying Tang and Yishi Piao and Yisong Wang and Yixuan Tan and Yiyang Ma and Yiyuan Liu and Yongqiang Guo and Yu Wu and Yuan Ou and Yuchen Zhu and Yuduan Wang and Yue Gong and Yuheng Zou and Yujia He and Yukun Zha and Yunfan Xiong and Yunxian Ma and Yuting Yan and Yuxiang Luo and Yuxiang You and Yuxuan Liu and Yuyang Zhou and Z. F. Wu and Z. Z. Ren and Zehui Ren and Zhangli Sha and Zhe Fu and Zhean Xu and Zhen Huang and Zhen Zhang and Zhenda Xie and Zhengyan Zhang and Zhewen Hao and Zhibin Gou and Zhicheng Ma and Zhigang Yan and Zhihong Shao and Zhipeng Xu and Zhiyu Wu and Zhongyu Zhang and Zhuoshu Li and Zihui Gu and Zijia Zhu and Zijun Liu and Zilin Li and Ziwei Xie and Ziyang Song and Ziyi Gao and Zizheng Pan},
      year={2025},
      eprint={2412.19437},
      archivePrefix={arXiv},
      primaryClass={cs.CL}, 
}

@inproceedings{ref34,
    title = "Justifying Recommendations using Distantly-Labeled Reviews and Fine-Grained Aspects",
    author = "Ni, Jianmo and Li, Jiacheng and McAuley, Julian",
    editor = "Inui, Kentaro and Jiang, Jing and Ng, Vincent and Wan, Xiaojun",
    booktitle = "Proceedings of the 2019 Conference on Empirical Methods in Natural Language Processing and the 9th International Joint Conference on Natural Language Processing (EMNLP-IJCNLP)",
    month = nov,
    year = "2019",
    address = "Hong Kong, China",
    publisher = "Association for Computational Linguistics",
    pages = "188--197"
}

@inproceedings{ref35,
author = {Rayana, Shebuti and Akoglu, Leman},
title = {Collective Opinion Spam Detection: Bridging Review Networks and Metadata},
year = {2015},
isbn = {9781450336642},
publisher = {Association for Computing Machinery},
address = {New York, NY, USA},
booktitle = {Proceedings of the 21th ACM SIGKDD International Conference on Knowledge Discovery and Data Mining},
pages = {985–994},
numpages = {10},
location = {Sydney, NSW, Australia},
series = {KDD '15}
}

@inproceedings{ref36,
 author = {Huang, Xuanwen and Yang, Yang and Wang, Yang and Wang, Chunping and Zhang, Zhisheng and Xu, Jiarong and Chen, Lei and Vazirgiannis, Michalis},
 booktitle = {Advances in Neural Information Processing Systems},
 editor = {S. Koyejo and S. Mohamed and A. Agarwal and D. Belgrave and K. Cho and A. Oh},
 pages = {22765--22777},
 publisher = {Curran Associates, Inc.},
 title = {DGraph: A Large-Scale Financial Dataset for Graph Anomaly Detection},
 volume = {35},
 year = {2022}
}

@inproceedings{ref38,
    title = "{MM}-{LLM}s: Recent Advances in {M}ulti{M}odal Large Language Models",
    author = "Zhang, Duzhen and Yu, Yahan and Dong, Jiahua and Li, Chenxing and Su, Dan and Chu, Chenhui and Yu, Dong",
    editor = "Ku, Lun-Wei  and
      Martins, Andre  and
      Srikumar, Vivek",
    booktitle = "Findings of the Association for Computational Linguistics: ACL 2024",
    month = aug,
    year = "2024",
    address = "Bangkok, Thailand",
    publisher = "Association for Computational Linguistics",
    pages = "12401--12430",
}

@inproceedings{ref39,
    title = "The Revolution of Multimodal Large Language Models: A Survey",
    author = "Caffagni, Davide and Cocchi, Federico and Barsellotti, Luca and Moratelli, Nicholas and Sarto, Sara and Baraldi, Lorenzo and Baraldi, Lorenzo and Cornia, Marcella  and Cucchiara, Rita",
    editor = "Ku, Lun-Wei and Martins, Andre and Srikumar, Vivek",
    booktitle = "Findings of the Association for Computational Linguistics: ACL 2024",
    month = aug,
    year = "2024",
    address = "Bangkok, Thailand",
    publisher = "Association for Computational Linguistics",
    pages = "13590--13618"
}

@article{ref50,
author = {Chang, Yupeng and Wang, Xu and Wang, Jindong and Wu, Yuan and Yang, Linyi and Zhu, Kaijie and Chen, Hao and Yi, Xiaoyuan and Wang, Cunxiang and Wang, Yidong and Ye, Wei and Zhang, Yue and Chang, Yi and Yu, Philip S. and Yang, Qiang and Xie, Xing},
title = {A Survey on Evaluation of Large Language Models},
year = {2024},
issue_date = {June 2024},
publisher = {Association for Computing Machinery},
address = {New York, NY, USA},
volume = {15},
number = {3},
issn = {2157-6904},
journal = {ACM Trans. Intell. Syst. Technol.},
month = mar,
articleno = {39},
numpages = {45}
}

@inproceedings{ref40,
author = {Kojima, Takeshi and Gu, Shixiang Shane and Reid, Machel and Matsuo, Yutaka and Iwasawa, Yusuke},
title = {Large language models are zero-shot reasoners},
year = {2022},
isbn = {9781713871088},
publisher = {Curran Associates Inc.},
address = {Red Hook, NY, USA},
booktitle = {Proceedings of the 36th International Conference on Neural Information Processing Systems},
articleno = {1613},
numpages = {15},
location = {New Orleans, LA, USA},
series = {NIPS '22}
}

@inproceedings{ref41,
title={Ano{LLM}: Large Language Models for Tabular Anomaly Detection},
author={Che-Ping Tsai and Ganyu Teng and Phillip Wallis and Wei Ding},
booktitle={The Thirteenth International Conference on Learning Representations},
year={2025}
}

@inproceedings{ref42,
title={ReTab{AD}: A Benchmark for Restoring Semantic Context in Tabular Anomaly Detection},
author={Sanghyu Yoon and Dongmin Kim and Suhee Yoon and Ye Seul Sim and Seungdong Yoa and Hye-Seung Cho and Soonyoung Lee and Hankook Lee and Woohyung Lim},
booktitle={The Fourteenth International Conference on Learning Representations},
year={2026}
}

@inproceedings{ref43,
  title={Tabula: Harnessing language models for tabular data synthesis},
  author={Zhao, Zilong and Birke, Robert and Chen, Lydia Y},
  booktitle={Pacific-Asia Conference on Knowledge Discovery and Data Mining},
  pages={247--259},
  year={2025},
  organization={Springer}
}

@inproceedings{ref44,
author = {Yeh, Chin-Chia Michael and Singh Saini, Uday and Dai, Xin and Fan, Xiran and Jain, Shubham and Fan, Yujie and Sun, Jiarui and Wang, Junpeng and Pan, Menghai and Dou, Yingtong and Chen, Yuzhong and Rakesh, Vineeth and Wang, Liang and Zheng, Yan and Das, Mahashweta},
title = {TREASURE: A Transformer-Based Foundation Model for High-Volume Transaction Understanding},
year = {2026},
isbn = {9798400722585},
publisher = {Association for Computing Machinery},
address = {New York, NY, USA},
booktitle = {Proceedings of the 32nd ACM SIGKDD Conference on Knowledge Discovery and Data Mining V.1},
pages = {2518–2527},
numpages = {10},
location = {Republic of Korea},
series = {KDD '26}
}

@inproceedings{ref45,
    title = "Enhancing Foundation Models in Transaction Understanding with {LLM}-based Sentence Embeddings",
    author = "Fan, Xiran and Jiang, Zhimeng and Yeh, Chin-Chia Michael and Chen, Yuzhong and Dou, Yingtong and Pan, Menghai and Zheng, Yan",
    editor = "Potdar, Saloni  and
      Rojas-Barahona, Lina  and
      Montella, Sebastien",
    booktitle = "Proceedings of the 2025 Conference on Empirical Methods in Natural Language Processing: Industry Track",
    month = nov,
    year = "2025",
    address = "Suzhou (China)",
    publisher = "Association for Computational Linguistics",
    pages = "903--911",
    ISBN = "979-8-89176-333-3"
}

@inproceedings{ref46,
author = {Dou, Yingtong and Jiang, Zhimeng and Zhang, Tianyi and Hu, Mingzhi and Xu, Zhichao and Chen, Huiyuan and Jain, Shubham and Saini, Uday Singh and Fan, Xiran and Sun, Jiarui and Pan, Menghai and Wang, Junpeng and Yeh, Chin-Chia Michael and Dai, Xin and Chen, Yuzhong},
title = {TransactionGPT: Toward Foundational Transaction Modeling},
year = {2026},
isbn = {9798400722592},
publisher = {Association for Computing Machinery},
address = {New York, NY, USA},
booktitle = {Proceedings of the 32nd ACM SIGKDD Conference on Knowledge Discovery and Data Mining V.2},
pages = {7185–7196},
numpages = {12},
location = {Republic of Korea},
series = {KDD '26}
}

@inproceedings{ref47,
author = {Zhang, Siwei and Xiong, Yun and Chen, Xi and Tang, Yateng and Jia, Zi'an and Zheng, Xuehao and Xu, Jiarong},
title = {Think-like-LSTM: Memory-Augmented Large Language Models via Dynamic Fine-Tuning for Financial Risk Assessment},
year = {2026},
isbn = {9798400722592},
publisher = {Association for Computing Machinery},
address = {New York, NY, USA},
booktitle = {Proceedings of the 32nd ACM SIGKDD Conference on Knowledge Discovery and Data Mining V.2},
pages = {8510–8521},
numpages = {12},
location = {Republic of Korea},
series = {KDD '26}
}

@inproceedings{ref48,
title={{PANTHER}: Generative Pretraining Beyond Language for Sequential User Behavior Modeling},
author={Guilin Li and Yun Zhang and Xiuyuan Chen and Chengqi Li and Bo Wang and Linghe Kong and Wenjia Wang and Weiran Huang and Matthias Hwai Yong Tan},
booktitle={The Thirty-ninth Annual Conference on Neural Information Processing Systems},
year={2025}
}

@inproceedings{ref49,
author = {Zhang, Zhongjian and Zhang, Mengmei and Xu, Dehua and Shi, Rongjun and Liu, Jianfeng and Meng, Fuli and Xu, Huajian and Wang, Xiao and Wang, Ruijia and Chen, Junze and Tang, Minwei and Shi, Chuan},
title = {FRiskGPT: A Generative Foundation Model for Financial Risk Detection},
year = {2026},
isbn = {9798400723070},
publisher = {Association for Computing Machinery},
address = {New York, NY, USA},
booktitle = {Proceedings of the ACM Web Conference 2026},
pages = {7733–7744},
numpages = {12},
location = {United Arab Emirates},
series = {WWW '26}
}

@ARTICLE{ref53,
  author={Han, Li and Wang, Longxun and Cheng, Ziyang and Wang, Bo and Yang, Guang and Cheng, Dawei and Lin, Xuemin},
  journal={IEEE Transactions on Knowledge and Data Engineering}, 
  title={Mitigating the Tail Effect in Fraud Detection by Community Enhanced Multi-Relation Graph Neural Networks}, 
  year={2025},
  volume={37},
  number={4},
  pages={2029-2041}}

@article{ref54,
    title={Sato: Contextual Semantic Type Detection in Tables},
    author={Dan Zhang and 
            Yoshihiko Suhara and 
            Jinfeng Li and 
            Madelon Hulsebos and 
            {\c{C}}a{\u{g}}atay Demiralp and 
            Wang-Chiew Tan},
    year = {2020},
    volume = {13},
    number = {12},
    journal = {Proc. VLDB Endow.},
    pages = {1835–1848},
    numpages = {14}
}

@inproceedings{ref55,
    title = "Logical Natural Language Generation from Open-Domain Tables",
    author = "Chen, Wenhu and Chen, Jianshu and Su, Yu and Chen, Zhiyu and Wang, William Yang",
    editor = "Jurafsky, Dan and Chai, Joyce and Schluter, Natalie and Tetreault, Joel",
    booktitle = "Proceedings of the 58th Annual Meeting of the Association for Computational Linguistics",
    month = jul,
    year = "2020",
    address = "Online",
    publisher = "Association for Computational Linguistics",
    pages = "7929--7942"
}

@inproceedings{ref56,
    title = "{PLOG}: Table-to-Logic Pretraining for Logical Table-to-Text Generation",
    author = "Liu, Ao and Dong, Haoyu and Okazaki, Naoaki and Han, Shi and Zhang, Dongmei",
    editor = "Goldberg, Yoav and Kozareva, Zornitsa and Zhang, Yue",
    booktitle = "Proceedings of the 2022 Conference on Empirical Methods in Natural Language Processing",
    month = dec,
    year = "2022",
    address = "Abu Dhabi, United Arab Emirates",
    publisher = "Association for Computational Linguistics",
    pages = "5531--5546"
}

@inproceedings{ref57,
    title = "{D}ata{N}arrative: Automated Data-Driven Storytelling with Visualizations and Texts",
    author = "Islam, Mohammed Saidul and Laskar, Md Tahmid Rahman and Parvez, Md Rizwan and Hoque, Enamul  and Joty, Shafiq",
    editor = "Al-Onaizan, Yaser and Bansal, Mohit and Chen, Yun-Nung",
    booktitle = "Proceedings of the 2024 Conference on Empirical Methods in Natural Language Processing",
    month = nov,
    year = "2024",
    address = "Miami, Florida, USA",
    publisher = "Association for Computational Linguistics",
    pages = "19253--19286"
}

@inproceedings{ref58,
  title={Mixture of knowledge minigraph agents for literature review generation},
  author={Zhang, Zhi and Liu, Yan and Zhong, Sheng-hua and Chen, Gong and Yang, Yu and Cao, Jiannong},
  booktitle={Proceedings of the AAAI Conference on Artificial Intelligence},
  volume={39},
  pages={26012--26020},
  year={2025}
}

@inproceedings{ref59,
  title={Learning to Generate and Extract: A Multi-Agent Collaboration Framework For Zero-shot Document-level Event Arguments Extraction},
  author={Zhang, Guangjun and Zhang, Hu and Han, Yazhou and Fan, Yue and Shao, Yuhang and Tan, Hongye and Li, Ru},
  booktitle={Proceedings of the AAAI Conference on Artificial Intelligence},
  volume={40},
  pages={34665--34673},
  year={2026}
}

@inproceedings{ref60,
    title = "{C}ontrol{M}ath: Controllable Data Generation Promotes Math Generalist Models",
    author = "Chen, Nuo and Wu, Ning and Chang, Jianhui and Shou, Linjun and Li, Jia",
    editor = "Al-Onaizan, Yaser and Bansal, Mohit and Chen, Yun-Nung",
    booktitle = "Proceedings of the 2024 Conference on Empirical Methods in Natural Language Processing",
    month = nov,
    year = "2024",
    address = "Miami, Florida, USA",
    publisher = "Association for Computational Linguistics",
    pages = "12201--12217"
}
\bibliographystyle{iclr2027_conference}

\newpage

\appendix

\section{External Knowledge and Semantic Priors}
\label{app:external_knowledge}

The semantic enrichment framework relies on external knowledge to ensure that generated attributes are not only semantically plausible but also statistically grounded and reproducible. We therefore prioritize authoritative statistical resources and domain-specific documentation that provide explicit distributions, category systems, or behavioral reference points relevant to the generated semantics. These resources are selected according to four considerations: coverage of the required semantic attributes, sufficient granularity for regional or category-level assignment, geographic alignment with the original data context of S-FFSD, and temporal alignment with the observation period of MS-FFSD. Geographic alignment ensures that the semantic priors reflect the same underlying population and market context as the transaction data. Since the transaction records span January to October 2021, we prioritize authoritative resources that are temporally close to the observation period and use the nearest available official statistics for each semantic component, rather than simply adopting the most recent data. This reduces temporal mismatch between the external priors and the transaction behavior that grounds semantic enrichment. The selected knowledge is subsequently transformed into structured distributions, constraints, or semantic priors before being incorporated into the corresponding enrichment components.

The selected resources serve distinct roles in enrichment. For temporal semantics, \textit{IEEE-CIS} provides a public reference for coarse temporal rhythms. The \textit{2020 Population Census} provides the primary regional population and demographic statistics for geographic assignment and user attributes such as gender, age, and education. The \textit{Statistical Yearbook 2021} supplements these with consumption expenditure, employment structure, and detailed age statistics for consumption, occupation, and age-related priors. On the merchant side, the \textit{Economic Census Yearbook 2023} informs merchant-count priors, while the Alipay merchant category documentation provides the hierarchical taxonomy for merchant categories. The \textit{2021 Economic and Social Development Statistical Bulletin} and the \textit{PwC 2021 Global Consumer Insights Survey} support category-level merchant profile priors. The online-tendency component is additionally calibrated using aggregate online-retail statistics reported in the \textit{2024 Economic and Social Development Statistical Bulletin} as a supplementary reference. Together, these sources provide complementary statistical, categorical, and behavioral priors at the granularity required by different enrichment components.

The primary external resources used in the framework are listed below:

\begin{itemize}
    \item \textit{IEEE-CIS}: 
    \url{https://www.kaggle.com/c/ieee-fraud-detection/}

    \item \textit{2020 Population Census}: 
    \url{https://www.stats.gov.cn/sj/pcsj/rkpc/d7c/202111/P020211126523667366751.pdf}

    \item \textit{Statistical Yearbook 2021}: 
    \url{https://www.stats.gov.cn/sj/ndsj/2021/indexch.htm}

    \item \textit{Economic Census Yearbook 2023}: 
    \url{https://www.stats.gov.cn/sj/pcsj/jjpc/5jp/zk/indexch.htm}

    \item \textit{Alipay Merchant Category Documentation}: 
    \url{https://opendocs.alipay.com/solution/0df0ir?pathHash=aea1536c}

    \item \textit{2021 Economic and Social Development Statistical Bulletin}: 
    \url{https://www.stats.gov.cn/xxgk/sjfb/zxfb2020/202202/t20220228_1827971.html}

    \item \textit{PwC 2021 Global Consumer Insights Survey}: 
    \url{https://runwise.co/wp-content/uploads/2021/12/%E6%B6%88%E8%B4%B9%E8%80%85%E6%B4%9E%E5%AF%9F%E6%8A%A5%E5%91%8A.pdf}
\end{itemize}

The external resources are used at different levels of granularity according to the semantic component being constructed. For user-related semantics, geographic assignment is constrained by regional population and consumption statistics, while gender, age group, education, and occupation industry are generated from demographic and employment distributions at the corresponding regional or attribute-specific levels. In particular, age-group assignment combines gender-specific fine-grained age statistics with regional broad-age distributions to construct a region-adjusted categorical prior.

For merchant-related semantics, consumption statistics provide the basis for broad consumption-category allocation, while business census information supplies an additional prior on the relative prevalence of merchant activities. The Alipay merchant category documentation defines the hierarchical candidate space for MCC Level 1 and MCC Level 2 assignment. Behavioral references from statistical bulletins and consumer research are further transformed into category-level merchant profile priors, and more recent online retail statistics are used to calibrate online tendency. Table~\ref{tab:external_knowledge} summarizes the primary external knowledge used in each component.

\begin{table}[ht]
\centering
\small
\caption{Primary external knowledge sources and their roles in semantic enrichment.}
\label{tab:external_knowledge}
\setlength{\tabcolsep}{3.5pt}
\renewcommand{\arraystretch}{1.0}

\begin{tabular}{
    >{\raggedright\arraybackslash}p{0.20\linewidth}
    >{\raggedright\arraybackslash}p{0.27\linewidth}
    >{\raggedright\arraybackslash}p{0.22\linewidth}
    >{\raggedright\arraybackslash}p{0.23\linewidth}
}
\toprule
\textbf{Component} &
\textbf{External Knowledge} &
\textbf{Granularity} &
\textbf{Usage} \\
\midrule

\rowcolor{my_purple}
\multicolumn{4}{c}{Temporal Priors} \\
\midrule

Temporal Rhythm
&
IEEE-CIS Fraud Detection
&
Coarse Temporal Activity Pattern
&
Temporal rhythm reference \\

\midrule

\rowcolor{my_blue}
\multicolumn{4}{c}{User-related Priors} \\
\midrule

Geographic Assignment
&
2020 Population Census
&
Region
&
User-count distribution \\

\midrule

Geographic Assignment
&
Statistical Yearbook 2021
&
Region
&
Transaction-amount distribution \\

\midrule

Gender
&
2020 Population Census
&
Region $\times$ Gender
&
Region-specific sampling \\

\midrule

Age Group
&
Statistical Yearbook 2021 + 2020 Population Census
&
Gender $\times$ 5-year Age Group; Region $\times$ Broad Age Group
&
Base age distribution and regional aging adjustment \\

\midrule

Education
&
2020 Population Census
&
Region $\times$ Education
&
Region-specific sampling \\

\midrule

Occupation Industry
&
Statistical Yearbook 2021
&
Region $\times$ Industry
&
Region-specific sampling \\

\midrule

\rowcolor{my_green}
\multicolumn{4}{c}{Merchant-related Priors} \\
\midrule

Consumption Category
&
Statistical Yearbook 2021
&
Consumption Category
&
Merchant amount prior \\

\midrule

Merchant Distribution
&
Economic Census Yearbook 2023
&
Industry
&
Merchant count prior \\

\midrule

Merchant Category
&
Alipay Merchant Category Documentation
&
MCC Level 1 $\times$ MCC Level 2
&
Merchant category space \\

\midrule

Merchant Profile
&
2021 Statistical Bulletin + PwC 2021 Survey
&
Category $\times$ Behavioral Attribute
&
MCC Level 2 profile priors \\

\bottomrule
\end{tabular}
\end{table}

\section{Mathematical Details of the Semantic Enrichment Framework}
\label{app:framework_details}

This section provides the detailed mathematical formulation of the multi-agent semantic enrichment framework, including transaction representation, agent dependencies, agent-specific objectives, constrained refinement, and textual generation.

\subsection{Problem Definition and Agent Workflow}

Let the anonymized transaction set be

\begin{equation}
\mathcal D=\{x_n\}_{n=1}^{N},\qquad
x_n=(o_n,u_n,m_n,a_n,l_n,c_n,y_n),
\label{eq:transaction_definition}
\end{equation}

where $o_n$ denotes the original transaction order, $u_n\in\mathcal U$ and $m_n\in\mathcal M$ denote the associated user and merchant, $a_n$ is the transaction amount, $l_n$ and $c_n$ are the anonymized location and transaction type, and $y_n$ is the original fraud label. The enrichment process introduces temporal, user, merchant, and textual semantics without overwriting these original fields.

The agent-level workflow is represented as a directed graph

\begin{equation}
\mathcal G=(\mathcal V,\mathcal E),\qquad
\mathcal V=\{A_T,A_U,A_M,A_C,A_X\},
\label{eq:agent_workflow}
\end{equation}

where $A_T$, $A_U$, $A_M$, $A_C$, and $A_X$ denote the Temporal, User, Merchant, Consistency, and Textual Agents, respectively. The dependency edges are

\begin{equation}
\begin{aligned}
\mathcal E=\{&
(A_T,A_M),(A_U,A_M),
(A_T,A_C),(A_U,A_C),\\
&(A_M,A_C),(A_T,A_X),(A_C,A_X)\}.
\end{aligned}
\label{eq:agent_dependencies}
\end{equation}

An edge $(A_i,A_j)\in\mathcal E$ indicates that the state or statistics produced by $A_i$ are used by $A_j$. Temporal and user states support merchant construction; temporal, user, and merchant states support cross-entity refinement; and the resulting structured states are subsequently passed to the Textual Agent.

To formalize structure preservation, let $\mathcal D^{+}$ denote the enriched dataset and $\Pi_{\mathrm{orig}}$ the projection onto the original transaction fields. Semantic enrichment is required to satisfy

\begin{equation}
\Pi_{\mathrm{orig}}(\mathcal D^{+})=\mathcal D.
\label{eq:original_preservation}
\end{equation}

Thus, the transaction order, participating entities, amounts, original anonymized attributes, and fraud labels remain unchanged throughout enrichment. The components summarized in Eq.~\ref{eq:framework_objective} are realized through staged construction and constrained refinement following the workflow above, after which the Textual Agent performs the generation mapping defined in Eq.~\ref{eq:textual_generation}.

\subsection{Temporal Semantic Construction}

The Temporal Agent constructs synthetic timestamps using coarse periodic priors derived from IEEE-CIS while preserving the original transaction order. Since IEEE-CIS provides relative timestamps rather than absolute calendar dates, they are mapped to a fixed pseudo-calendar solely for estimating periodic temporal statistics.

The daily phase is first calibrated by aligning the lowest-activity window with a predefined early-morning interval. For each candidate hourly shift $\delta$, let $C_h^{(\delta)}$ denote the transaction count at hour $h$ after applying the shift, and define

\begin{equation}
s_\delta=
\arg\min_{s}
\sum_{j=0}^{L-1}
C_{(s+j)\bmod 24}^{(\delta)}.
\end{equation}

The selected shift is

\begin{equation}
\delta^\star=
\arg\min_{\delta}
\left[
\operatorname{cdist}(s_\delta,h_0),
\sum_{j=0}^{L-1}
C_{(s_\delta+j)\bmod 24}^{(\delta)}
\right]_{\mathrm{lex}},
\label{eq:temporal_phase}
\end{equation}

where $L$ is the low-activity window length, $h_0$ its target starting hour, $\operatorname{cdist}$ denotes circular distance, and $[\cdot]_{\mathrm{lex}}$ denotes lexicographic minimization. The shifted relative timestamps are then anchored to a fixed reference date to obtain a pseudo-calendar representation. 

From this representation, the agent estimates a half-hour-of-week distribution
$p_{\mathrm{how}}^{\mathrm{ref}}$, a day-of-month distribution
$p_{\mathrm{dom}}^{\mathrm{ref}}$, and a fraud-specific half-hour-of-week distribution
$p_{\mathrm{how}}^{\mathrm{fraud}}$. Each distribution is obtained from smoothed empirical counts,

\begin{equation}
p(j)=
\frac{C_j+\alpha}
{\sum_k C_k+\alpha|\mathcal B_{\mathrm{time}}|},
\label{eq:temporal_smoothed_prior}
\end{equation}

where $\mathcal B_{\mathrm{time}}$ denotes the corresponding temporal-bin space and $\alpha$ is the smoothing parameter. 

For users with sufficient transaction evidence, source-specific rhythms are constructed by matching their behavioral profiles to reference pseudo users. Let $\mathcal N_K(u)$ denote the matched reference users and $d_{uj}$ their profile distance. Their temporal distributions are aggregated as

\begin{equation}
\bar p_u^{\mathrm{ref}}
=
\sum_{j\in\mathcal N_K(u)}
\beta_{uj}p_j^{\mathrm{ref}},
\qquad
\beta_{uj}
=
\frac{\exp(-d_{uj})}
{\sum_{k\in\mathcal N_K(u)}\exp(-d_{uk})},
\end{equation}

and combined with the global rhythm as

\begin{equation}
p_u^{\mathrm{src}}
=
\eta_u\,\bar p_u^{\mathrm{ref}}
+
(1-\eta_u)\,p_{\mathrm{how}}^{\mathrm{ref}},
\label{eq:temporal_user_rhythm}
\end{equation}

where $\eta_u\in[0,1]$ controls the contribution of the matched user-specific rhythm.

The target observation interval is partitioned into calendar bins. For a bin $b$, its global weight combines the half-hour-of-week and day-of-month priors:

\begin{equation}
w_{\mathrm{global}}(b)
\propto
p_{\mathrm{how}}^{\mathrm{ref}}\!\left(h(b)\right)
p_{\mathrm{dom}}^{\mathrm{ref}}\!\left(d(b)\right),
\label{eq:temporal_global_weight}
\end{equation}

where $h(b)$ and $d(b)$ denote its half-hour-of-week and day-of-month indices. For each order-preserving transaction block $B$, the temporal distribution is defined as

\begin{equation}
p_B^{\mathrm{time}}
=
\begin{cases}
\eta_f p_{\mathrm{how}}^{\mathrm{fraud}}
+\eta_s p_u^{\mathrm{src}}
+\eta_g p_{\mathrm{how}}^{\mathrm{ref}},
& B\text{ is fraud-related},\\[1mm]
\eta_m p_u^{\mathrm{src}}
+(1-\eta_m)p_{\mathrm{how}}^{\mathrm{ref}},
& u\text{ is matched},\\[1mm]
p_{\mathrm{how}}^{\mathrm{ref}},
& \text{otherwise},
\end{cases}
\label{eq:temporal_block_rhythm}
\end{equation}

where $\eta_f,\eta_s,\eta_g\geq0$, $\eta_f+\eta_s+\eta_g=1$, and $\eta_m\in[0,1]$. Candidate start bins are then sampled according to

\begin{equation}
w_B(b)
\propto
\eta_b\,w_{\mathrm{global}}(b)
+
(1-\eta_b)\,
p_B^{\mathrm{time}}\!\left(h(b)\right),
\label{eq:temporal_block_weight}
\end{equation}

where $\eta_b\in[0,1]$ balances global calendar rhythms and block-specific temporal patterns. 

Fraud-related blocks additionally reuse relative inter-transaction intervals extracted from reference fraud bursts when a suitable template is available; otherwise compact intervals are generated. The resulting intervals are fitted within the admissible calendar segment while maintaining non-decreasing transaction order. Distributional discrepancies used by subsequent components are measured using total variation distance,

\begin{equation}
D_{\mathrm{TV}}(p,q)
=
\frac{1}{2}\sum_v |p(v)-q(v)|.
\label{eq:tv_distance}
\end{equation}

Finally, the generated timestamps satisfy

\begin{equation}
\tau_{(1)}\leq\tau_{(2)}\leq\cdots\leq\tau_{(N)},
\qquad
\tau_n\in[t_{\min},t_{\max}],
\label{eq:temporal_constraints}
\end{equation}

where $(1),\ldots,(N)$ follows the ordering induced by the original order field $o_n$. Thus, temporal enrichment introduces calendar semantics without altering the original transaction sequence.

\subsection{User Semantic Initialization}

The User Agent initializes geographic and demographic semantics from observed user--location interactions and external priors. Let
$B\in\mathbb{R}^{|\mathcal U|\times|\mathcal L|}$ denote the user--location interaction matrix, where

\begin{equation}
B_{u\ell}=
\sum_{n=1}^{N}
\mathbb I[u_n=u,\,l_n=\ell].
\label{eq:user_location_matrix}
\end{equation}

Locations exhibiting similar user interaction patterns are identified using cosine similarity,

\begin{equation}
s(\ell,\ell')=
\frac{B_{:\ell}^{\top}B_{:\ell'}}
{\|B_{:\ell}\|_2\|B_{:\ell'}\|_2},
\label{eq:location_similarity}
\end{equation}

and thresholded similarities are used to form a set of location blocks $\mathcal B$. Let
$\rho:\mathcal B\rightarrow\mathcal R$ denote the mapping from location blocks to geographic regions. The mapping jointly matches regional distributions measured by user count and transaction amount:

\begin{equation}
\rho^\star=
\arg\min_{\rho}
\lambda_N
D_{\mathrm{TV}}
\!\left(
\widehat p_{\rho}^{N},
p_{\mathrm{ref}}^{N}
\right)
+
\lambda_A
D_{\mathrm{TV}}
\!\left(
\widehat p_{\rho}^{A},
p_{\mathrm{ref}}^{A}
\right),
\label{eq:geographic_mapping}
\end{equation}

where $\widehat p_{\rho}^{N}$ and $\widehat p_{\rho}^{A}$ are the resulting geographic distributions by user count and transaction amount, respectively. These two terms correspond to $g_n$ and $g_a$ in Eq.~\ref{eq:initialization_objectives}.

After assigning the geographic attribute $r_u^{\mathrm{prov}}$, demographic attributes are sampled conditionally as

\begin{equation}
\begin{aligned}
r_u^{\mathrm{gender}} &\sim P(G\mid r_u^{\mathrm{prov}}),\\
r_u^{\mathrm{age}} &\sim P(A\mid r_u^{\mathrm{prov}},r_u^{\mathrm{gender}}),\\
r_u^{\mathrm{edu}} &\sim P(E\mid r_u^{\mathrm{prov}}),\\
r_u^{\mathrm{occ}} &\sim P(O\mid r_u^{\mathrm{prov}}).
\end{aligned}
\label{eq:demographic_sampling}
\end{equation}

Accordingly, the demographic component $d$ is expressed through conditional distribution matching. Expanding the User Agent objective gives

\begin{equation}
\begin{aligned}
F_U(\boldsymbol r)={}&
\lambda_N
D_{\mathrm{TV}}
(\widehat p_{\mathrm{prov}}^{N},
 p_{\mathrm{prov}}^{N,\mathrm{ref}})
+
\lambda_A
D_{\mathrm{TV}}
(\widehat p_{\mathrm{prov}}^{A},
 p_{\mathrm{prov}}^{A,\mathrm{ref}})\\
&+
\sum_{f\in\mathcal F_U}
\lambda_f
\sum_{\xi\in\Xi_f}
\pi_\xi
D_{\mathrm{TV}}
\!\left(
\widehat p_f(\cdot\mid\xi),
p_f^{\mathrm{ref}}(\cdot\mid\xi)
\right),
\end{aligned}
\label{eq:user_objective_detail}
\end{equation}

where
$\mathcal F_U=\{\mathrm{gender},\mathrm{age},\mathrm{education},\mathrm{occupation}\}$.
For gender, education, and occupation, $\Xi_f$ represents the geographic region used for conditioning, while age is conditioned jointly on geographic region and gender. $\pi_\xi$ denotes the weight of context $\xi$. Thus, geographic semantics are grounded in user--location behavior and regional distributions, while demographic semantics preserve the corresponding conditional priors.

\subsection{Merchant Semantic Initialization}

The Merchant Agent constructs hierarchical business semantics by combining observed transaction behavior with merchant-side priors. For each merchant $m$, we define a behavioral profile

\begin{equation}
\phi_m=
\left(
n_m,\bar a_m,\operatorname{cv}_m,
r_m^{\mathrm{cross}},
H_m,
P_m^{\mathrm{demo}}
\right),
\label{eq:merchant_profile}
\end{equation}

where $n_m$ is the transaction count, $\bar a_m$ the mean transaction amount, $\operatorname{cv}_m$ the amount coefficient of variation, $r_m^{\mathrm{cross}}$ the cross-region transaction ratio, $H_m$ the dominant activity periods, and $P_m^{\mathrm{demo}}$ the demographic distribution of associated users.

Let $c_m\in\mathcal C_M$ denote the broad consumption category assigned to merchant $m$. Category initialization targets the reference merchant-count and transaction-amount distributions, summarized by

\begin{equation}
\mathcal L_{\mathrm{broad}}^M(\boldsymbol c)
=
\lambda_c
D_{\mathrm{TV}}
\!\left(
\widehat p_c^{\mathrm{count}},
p_c^{\mathrm{count}}
\right)
+
\lambda_a
D_{\mathrm{TV}}
\!\left(
\widehat p_c^{\mathrm{amount}},
p_c^{\mathrm{amount}}
\right).
\label{eq:merchant_category_assignment}
\end{equation}

The assignment is implemented greedily by tracking category-level count and amount quotas. The two terms correspond to the merchant-count and transaction-amount components $c$ and $a$ in Eq.~\ref{eq:initialization_objectives}. Within each assigned broad category, finer merchant categories are initialized from the admissible hierarchy using category-relative behavioral semantics and soft distributional balancing. Accordingly, letting $q_m=(c_m,k_m)$ denote the structured merchant state, the Merchant Agent objective is summarized as

\begin{equation}
F_M(\boldsymbol q;\boldsymbol r,\boldsymbol{\tau})
=
\lambda_c D_{\mathrm{TV}}(\widehat p_c^{\mathrm{count}},p_c^{\mathrm{count}})
+
\lambda_a D_{\mathrm{TV}}(\widehat p_c^{\mathrm{amount}},p_c^{\mathrm{amount}})
+
\lambda_s\mathcal L_s^M(\boldsymbol q),
\label{eq:merchant_objective_detail}
\end{equation}

where $\mathcal L_s^M$ captures category-relative semantic compatibility and within-category balancing during hierarchical initialization, corresponding to the semantic component $s$ in Eq.~\ref{eq:initialization_objectives}.

For subsequent cross-entity refinement, the compatibility between merchant $m$ and candidate fine-grained category $k$ is measured by

\begin{equation}
\begin{aligned}
S_M(m,k)={}&
\alpha_o\!\left(
1-
\left|\widetilde r_m^{\mathrm{cross}}
-q_k^{\mathrm{online}}\right|
\right)
+\alpha_a\!\left(1-d_{\mathrm{ord}}(a_m,a_k)\right)\\
&+\alpha_v\!\left(1-d_{\mathrm{ord}}(v_m,v_k)\right)
+\alpha_t J(H_m,H_k)\\
&+\frac{\alpha_g}{|\mathcal F_D|}
\sum_{f\in\mathcal F_D}
\left[
1-D_{\mathrm{TV}}(P_m^f,Q_k^f)
\right]
+\alpha_s\log\omega_k .
\end{aligned}
\label{eq:merchant_semantic_score}
\end{equation}

Here, $\widetilde r_m^{\mathrm{cross}}$ denotes the normalized cross-region tendency derived from $r_m^{\mathrm{cross}}$, and $q_k^{\mathrm{online}}$ is the online tendency associated with category $k$. The ordinal variables $a_m$ and $v_m$ are derived from $\bar a_m$ and $\operatorname{cv}_m$, respectively, while $a_k$ and $v_k$ denote the corresponding category-level references. The normalized ordinal distance is

\begin{equation}
d_{\mathrm{ord}}(x,y)
=
\frac{
|\operatorname{rank}(x)-\operatorname{rank}(y)|
}{
|\mathcal L|-1
},
\label{eq:ordinal_distance}
\end{equation}

where $\mathcal L$ is the ordered set of semantic levels, $J(\cdot,\cdot)$ denotes Jaccard similarity, and $\mathcal F_D$ denotes the demographic attributes used for merchant--customer consistency. $P_m^f$ and $Q_k^f$ denote the observed and reference customer distributions for attribute $f$, respectively, $H_k$ denotes the reference active periods of category $k$, and $\omega_k$ is a balancing weight. This score is subsequently reused by the Consistency Agent to refine fine-grained merchant semantics.

\subsection{Cross-Entity Semantic Refinement}

The Consistency Agent refines initialized user and merchant semantics according to their observed interactions. The transaction records induce a bipartite interaction graph

\begin{equation}
\mathcal G_{UM}
=
(\mathcal U\cup\mathcal M,\mathcal E_{UM}),
\qquad
(u,m)\in\mathcal E_{UM}
\iff
\exists n:\,u_n=u,\ m_n=m,
\label{eq:user_merchant_graph}
\end{equation}

where an edge indicates that at least one transaction is observed between user $u$ and merchant $m$. This graph differs from the agent workflow graph $\mathcal G$: $\mathcal G_{UM}$ represents entity interactions, whereas $\mathcal G$ describes information dependencies among agents.

For a refinable user attribute
$f\in\mathcal F_R=\{\mathrm{age},\mathrm{education},\mathrm{occupation}\}$
and candidate value $v$, the user-side consistency score is defined as

\begin{equation}
\begin{aligned}
S_U(u,f,v)={}&
\gamma_p P_{\mathcal N(u)}^f(v)
+\gamma_m
\left[
1-D_{\mathrm{TV}}
(P_u^{\mathrm{MCC}},Q_v^{\mathrm{MCC}})
\right]\\
&+\gamma_a C_a(u,v)
+\gamma_t C_t(u,v)
+\gamma_s\log\omega_{f,v}.
\end{aligned}
\label{eq:user_consistency_score}
\end{equation}

Let $n_{um}$ denote the number of transactions between user $u$ and merchant $m$, and $P_m^f(v)$ the prevalence of attribute value $v$ among users associated with merchant $m$. The merchant-mediated prevalence is

\begin{equation}
P_{\mathcal N(u)}^f(v)
=
\frac{
\sum_{m\in\mathcal N(u)}
n_{um}P_m^f(v)
}{
\sum_{m\in\mathcal N(u)}n_{um}
}.
\label{eq:merchant_mediated_prevalence}
\end{equation}

$P_u^{\mathrm{MCC}}$ denotes the merchant-category distribution observed for user $u$, and $Q_v^{\mathrm{MCC}}$ the corresponding reference distribution for candidate value $v$. Amount and temporal compatibility are defined as

\begin{equation}
C_a(u,v)
=
1-d_{\mathrm{ord}}(a_u,a_v^{\mathrm{ref}}),
\qquad
C_t(u,v)
=
J(H_u,H_v^{\mathrm{ref}}),
\label{eq:user_behavior_compatibility}
\end{equation}

where $a_u$ and $H_u$ denote the user's ordinal amount level and dominant activity periods, and $a_v^{\mathrm{ref}}$ and $H_v^{\mathrm{ref}}$ are their reference counterparts associated with $v$. The balancing term $\omega_{f,v}$ discourages excessive concentration on particular semantic values.

Using $S_U$ together with the merchant compatibility score $S_M$ defined in Eq.~\ref{eq:merchant_semantic_score}, the two consistency components in Eq.~\ref{eq:consistency_objective} are

\begin{equation}
\mathcal L_U^C
=
-\sum_{u\in\mathcal U}
\sum_{f\in\mathcal F_R}
S_U(u,f,r_u^f),
\qquad
\mathcal L_M^C
=
-\sum_{m\in\mathcal M}
S_M(m,k_m).
\label{eq:cross_entity_consistency_terms}
\end{equation}

Thus, larger compatibility scores reduce the consistency objective, while the change penalty in Eq.~\ref{eq:semantic_gain} discourages unnecessary reassignment. User attributes are refined independently, and merchant candidates remain restricted to fine-grained categories within the parent category assigned during initialization.

At iteration $t$, the feasible candidate set for entity $e$ is denoted by $\mathcal A_e^{(t)}$. Using the net gain in Eq.~\ref{eq:semantic_gain}, the update rule is

\begin{equation}
z_e^{(t+1)}
=
\begin{cases}
\displaystyle
\arg\max_{z\in\mathcal A_e^{(t)}}
\Delta_e(z;Z^{(t)}),
&
\displaystyle
\max_{z\in\mathcal A_e^{(t)}}
\Delta_e(z;Z^{(t)})>0,\\[4pt]
z_e^{(t)},
&
\text{otherwise}.
\end{cases}
\label{eq:positive_gain_update}
\end{equation}

For constrained user attributes, the Constraint Engine adjusts candidate weights as

\begin{equation}
\omega_{f,v}^{(t)}
=
\operatorname{clip}
\left(
1+\eta
\frac{
p_{f,v}^{\mathrm{ref}}
-\widehat p_{f,v}^{(t)}
}{
b_{f,v}
},
\omega_{\min},
\omega_{\max}
\right),
\qquad
b_{f,v}
=
\max\left(
p_{f,v}^{\mathrm{ref}},
\frac{1}{|\mathcal V_f|}
\right),
\label{eq:constraint_weight}
\end{equation}

where $\mathcal V_f$ is the value space of attribute $f$. Underrepresented values therefore receive larger weights and overrepresented values receive smaller ones. Final user distributions satisfy

\begin{equation}
D_{\mathrm{TV}}
\left(
\widehat p_f,p_f^{\mathrm{ref}}
\right)
\leq \epsilon_f,
\qquad
f\in\mathcal F_R.
\label{eq:distribution_tolerance}
\end{equation}

Merchant-side refinement is additionally constrained by category-level structural limits on modification and semantic concentration.

Let $\Delta\mathcal U^{(t)}$ and $\Delta\mathcal M^{(t)}$ denote the user and merchant sets changed during the main update stage. Only directly affected counterparts are activated for response:

\begin{equation}
\begin{aligned}
\mathcal R_M^{(t)}
&=
\{m\in\mathcal M:
\exists u\in\Delta\mathcal U^{(t)},
(u,m)\in\mathcal E_{UM}\},\\
\mathcal R_U^{(t)}
&=
\{u\in\mathcal U:
\exists m\in\Delta\mathcal M^{(t)},
(u,m)\in\mathcal E_{UM}\}.
\end{aligned}
\label{eq:local_response_sets}
\end{equation}

Responses are therefore localized to entities whose connected counterparts have changed, rather than rescanning the full user or merchant population. Response updates are bounded and do not recursively trigger unbounded cascades.

Let $N_U^{(t)}$ and $N_M^{(t)}$ denote the numbers of accepted primary-stage user-field and merchant-category updates at iteration $t$, respectively, and let $N_U$ and $N_M$ denote the corresponding candidate-pool sizes. The primary-stage modification budgets satisfy

\begin{equation}
\begin{aligned}
N_U^{(t)}
&\leq b_U^{\mathrm{round}}N_U,
&
\sum_t N_U^{(t)}
&\leq b_U^{\mathrm{total}}N_U,\\
N_M^{(t)}
&\leq b_M^{\mathrm{round}}N_M,
&
\sum_t N_M^{(t)}
&\leq b_M^{\mathrm{total}}N_M.
\end{aligned}
\label{eq:refinement_budgets}
\end{equation}

The refinement terminates when no admissible positive-gain update remains, the modification budgets are exhausted, the accepted gains become sufficiently small for consecutive rounds, or the maximum number of iterations is reached. If a final hard constraint is violated, eligible low-gain updates associated with the violated attribute or category are reverted, with the constraints re-evaluated after each repair step.

\subsection{Textual Semantic Generation}

The Textual Agent first constructs an evidence-grounded representation for each entity from its refined structured semantics and observed transaction context. For users and merchants, the corresponding inputs are defined as

\begin{equation}
\begin{aligned}
h_u &=
\operatorname{Serialize}\!\left(
r_u,
\operatorname{Agg}\{\chi(x_n):u_n=u\}
\right),\\
h_m &=
\operatorname{Serialize}\!\left(
q_m,
\operatorname{Agg}\{\chi(x_n):m_n=m\},
\operatorname{Agg}\{r_{u_n}:m_n=m\}
\right),
\end{aligned}
\label{eq:textual_entity_inputs}
\end{equation}

where $\chi(x_n)$ extracts transaction-level temporal, amount, geographic, and interaction statistics, $\operatorname{Agg}$ summarizes information associated with the entity, and $\operatorname{Serialize}$ converts the resulting structured state into the input representation used for text generation.

Let
$n_u=\sum_{n=1}^{N}\mathbb I[u_n=u]$
denote the number of observed transactions associated with user $u$. User descriptions are generated according to the available transaction evidence:

\begin{equation}
d_u=
\begin{cases}
\Gamma(h_u), & n_u=1,\\
\operatorname{LLM}\!\left(P_U(h_u;\mathcal K)\right), & n_u>1,
\end{cases}
\label{eq:user_description_generation}
\end{equation}

where $\Gamma$ denotes the deterministic template used for single-transaction users, and $P_U$ denotes the structured user prompt. The deterministic template verbalizes the available user attributes together with the geographic, temporal, and merchant-category context of the observed transaction. Single-transaction users are handled deterministically because one observation does not provide sufficient evidence for reliable behavioral aggregation. For multi-transaction users, aggregated behavioral patterns are combined with the refined user semantics and provided to the fixed language model.

Merchant descriptions are generated as

\begin{equation}
d_m=
\operatorname{LLM}\!\left(
P_M(h_m;\mathcal K,\mathcal H_{c_m})
\right),
\label{eq:merchant_description_generation}
\end{equation}

where $P_M$ denotes the merchant prompt and $\mathcal H_{c_m}$ stores previously generated concepts within the same category to reduce repetition. Textual generation is conditioned on the refined semantic state and observed evidence without modifying the structured state. The complete prompts and deterministic templates are provided in Appendix~\ref{app:text_generation_prompts}.

\subsection{Shared Constraints and Control}

The agents are coordinated through a shared external knowledge base, a global feasible region, and fraud-aware preservation rules. The external knowledge base $\mathcal K$ contains the temporal rhythms, demographic distributions, consumption structures, merchant category systems, and behavioral priors used by the corresponding agents. The specific knowledge sources, statistical granularity, temporal alignment, and their roles are detailed in Appendix~\ref{app:external_knowledge}.

The Constraint Engine defines the feasible region $\mathcal C(\mathcal D,\mathcal K)$ by intersecting several constraint sets:
\begin{equation}
\mathcal C(\mathcal D,\mathcal K)
=
\mathcal C_{\mathrm{orig}}
\cap
\mathcal C_{\mathrm{time}}
\cap
\mathcal C_{\mathrm{dist}}
\cap
\mathcal C_{\mathrm{hier}}
\cap
\mathcal C_{\mathrm{budget}},
\label{eq:feasible_region}
\end{equation}

where $\mathcal C_{\mathrm{orig}}$ preserves the original transaction fields as specified in Eq.~\ref{eq:original_preservation}, $\mathcal C_{\mathrm{time}}$ enforces temporal ordering and valid timestamp ranges, $\mathcal C_{\mathrm{dist}}$ constrains deviations from reference distributions, $\mathcal C_{\mathrm{hier}}$ restricts semantic changes to valid category hierarchies, and $\mathcal C_{\mathrm{budget}}$ bounds the extent of semantic modification. Together, these constraint sets provide shared control across enrichment stages.

The Fraud-aware Preservation Controller further protects abnormal patterns supported by transaction evidence, including temporal bursts, geographic anomalies, and residual merchant abnormalities, from being removed solely to improve global semantic consistency. Together, the shared knowledge and constraints coordinate stage-specific updates while maintaining structural, distributional, and fraud-related preservation requirements.

\begin{figure}[t]
    \centering
    \includegraphics[width=\linewidth]{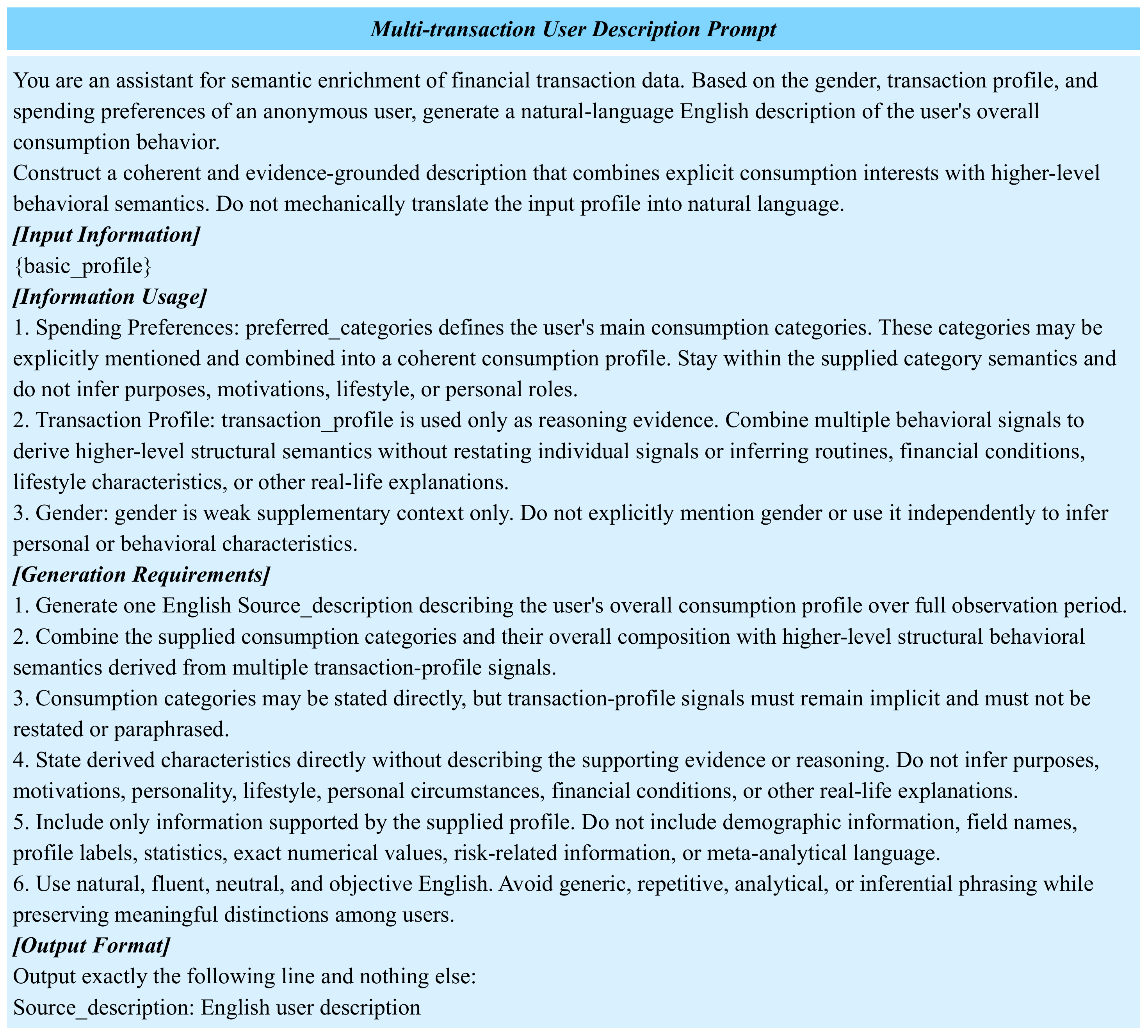}
    \caption{Prompt used for multi-transaction user description generation.}
    \label{fig:user_description_prompt}
\end{figure}

\begin{figure}[!ht]
    \centering
    \includegraphics[width=\linewidth]{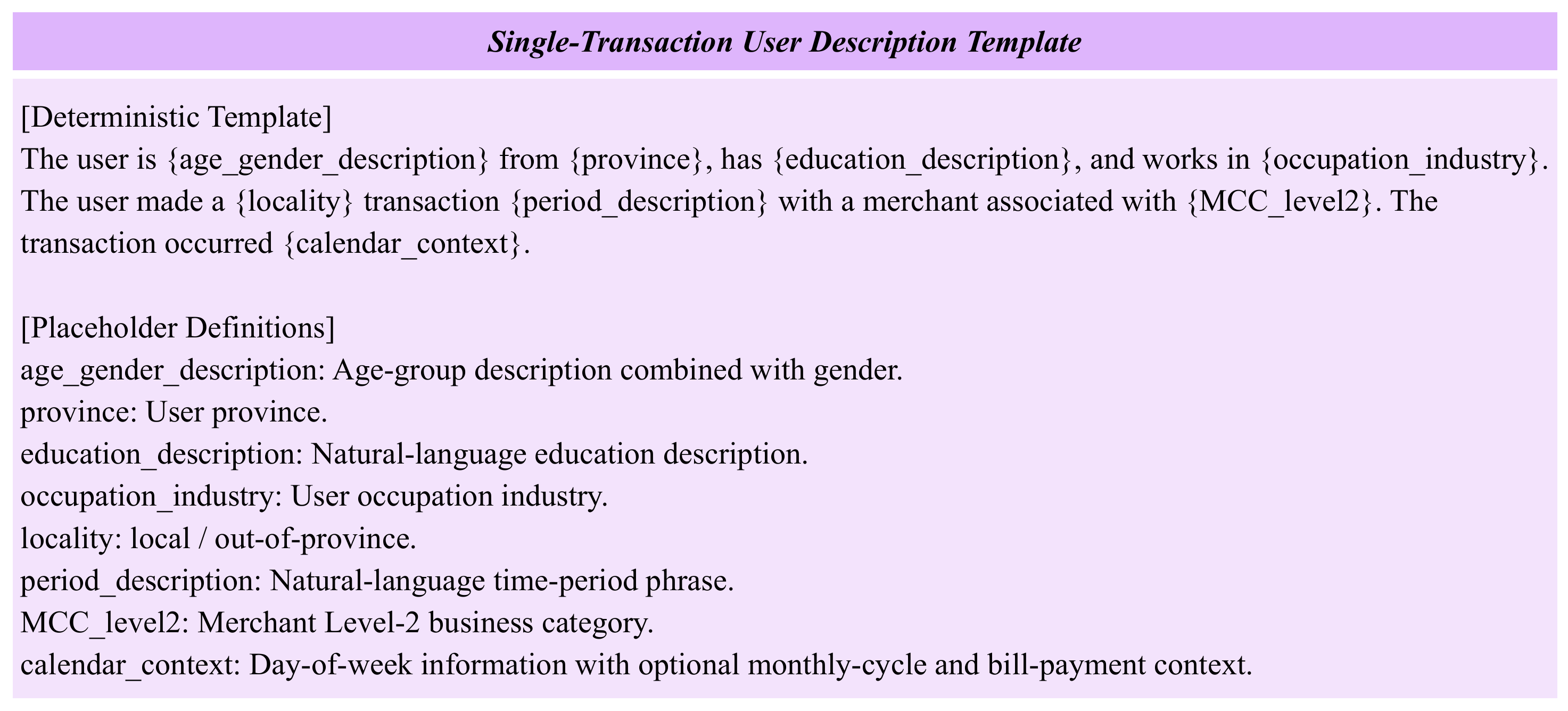}
    \caption{Deterministic template used for single-transaction user description generation.}
    \label{fig:single_transaction_user_template}
\end{figure}

\begin{figure}[t]
    \centering
    \includegraphics[width=\linewidth]{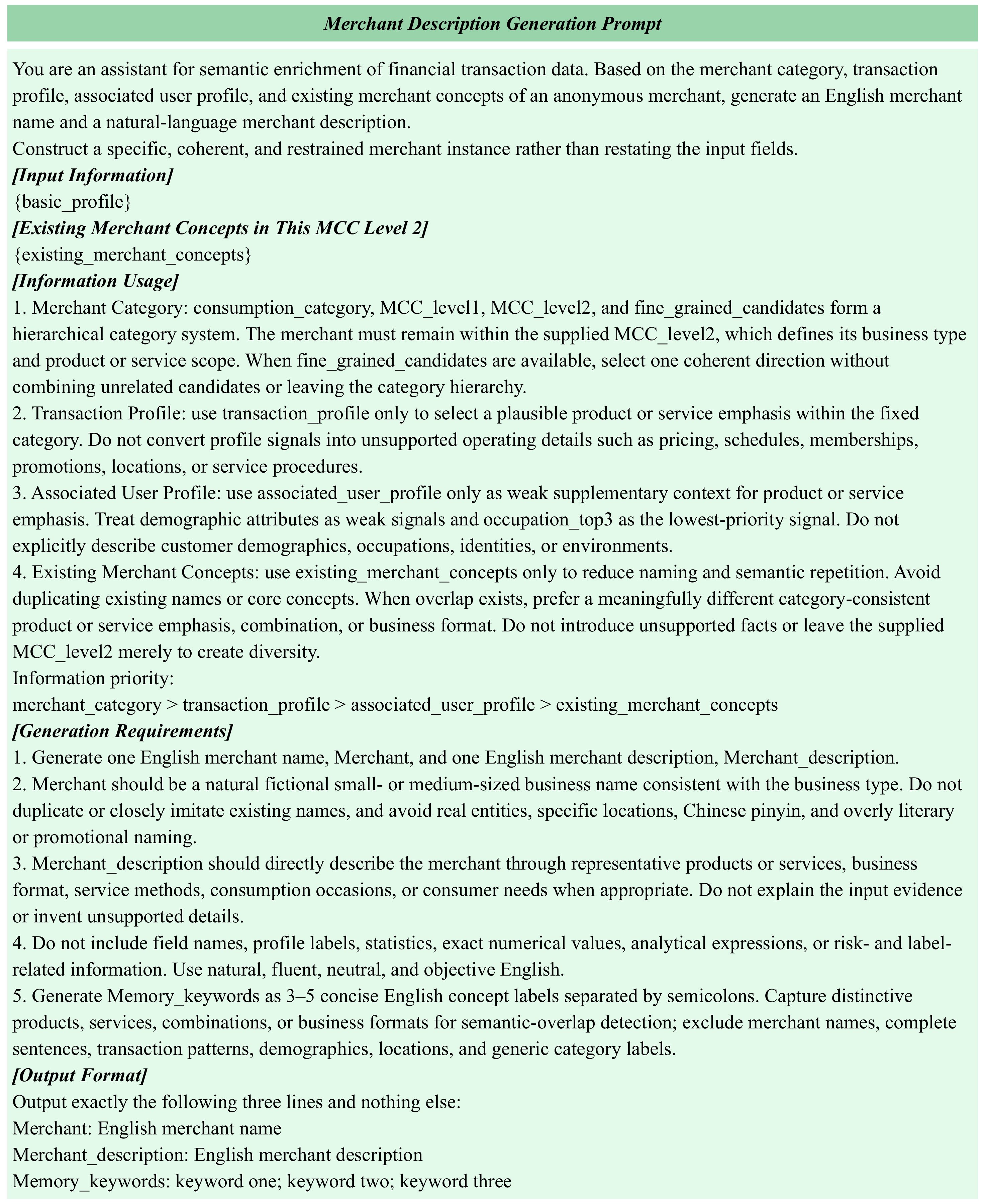}
    \caption{Prompt used for merchant description generation.}
    \label{fig:merchant_description_prompt}
\end{figure}

\section{Textual Semantic Generation Prompts}
\label{app:text_generation_prompts}

\subsection{User Description Generation}

For users with multiple observed transactions, we first aggregate their transaction history into a structured behavioral profile containing transaction patterns and spending preferences. The generation input retains only the information required for textual construction, and the prompt instructs the language model to combine explicit consumption categories with higher-level behavioral semantics derived from multiple transaction signals. Individual transaction-profile fields are not directly reproduced in the description, and demographic information is not explicitly expressed. The complete prompt is shown in Figure~\ref{fig:user_description_prompt}.

For users with only one observed transaction, behavioral aggregation is not sufficiently supported by repeated evidence. We therefore use a deterministic template rather than LLM-based generation. The template verbalizes the transaction’s geographic, temporal, and merchant-category context, producing a concise description directly from the observed and enriched fields. The template and its placeholder definitions are shown in Figure~\ref{fig:single_transaction_user_template}.

The template verbalizes the transaction’s geographic, temporal, and merchant-category context without repeating demographic attributes already represented in the structured user profile.

\subsection{Merchant Description Generation}

\begin{figure}[t]
    \centering
    \includegraphics[width=\linewidth]{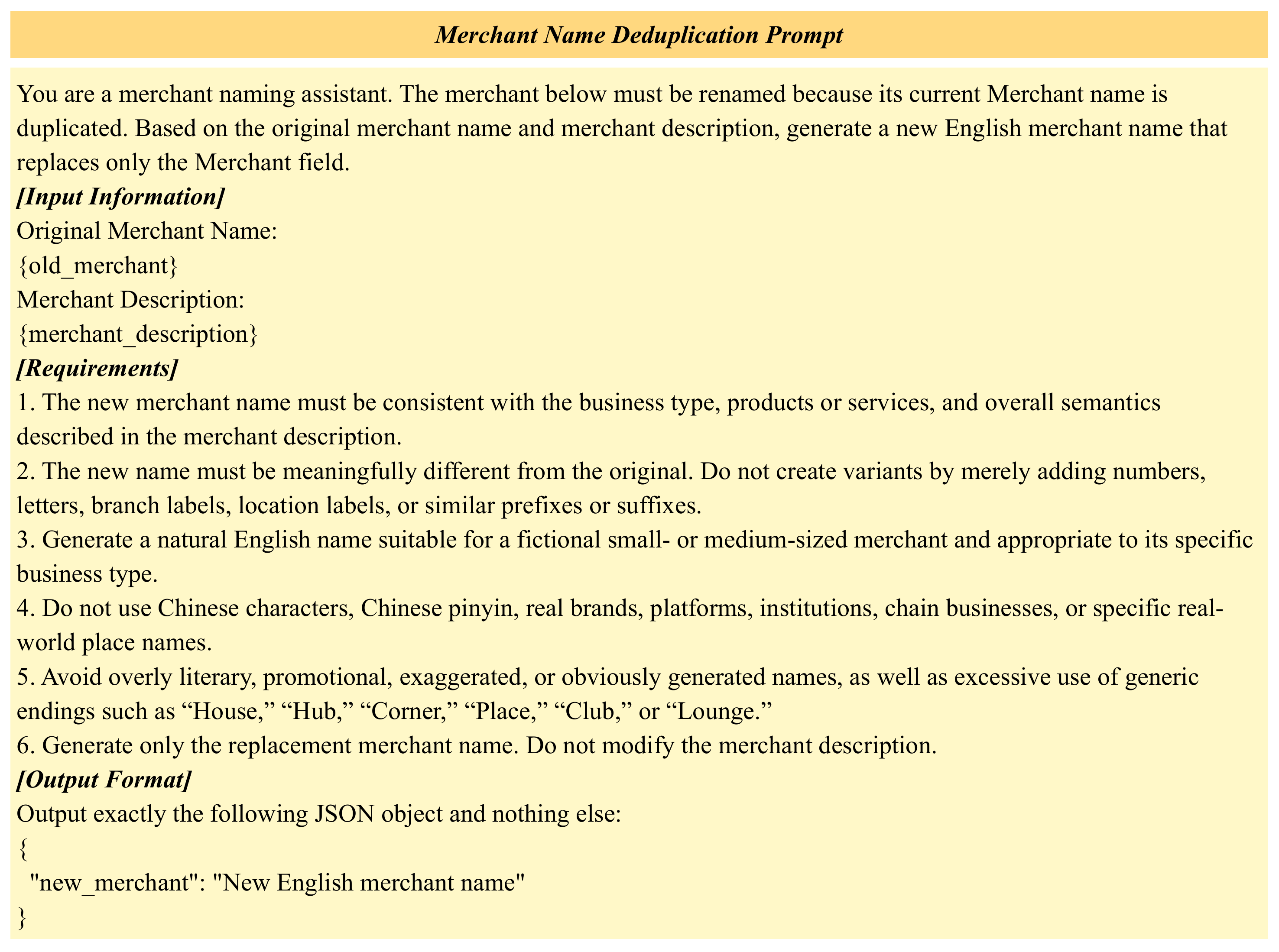}
    \caption{Prompt used for duplicate merchant name resolution.}
    \label{fig:merchant_name_deduplication_prompt}
\end{figure}

Merchant descriptions are generated from the merchant category hierarchy, transaction profile, and aggregated profiles of associated users. The category hierarchy defines the admissible business scope, while transaction and user information provide supporting context for selecting a plausible product or service emphasis. The model is instructed to construct a specific and coherent merchant instance without directly reproducing profile fields or introducing unsupported operating details. The complete generation prompt is shown in Figure~\ref{fig:merchant_description_prompt}.

To reduce semantic repetition among merchants within the same MCC Level 2 category, the generation process maintains category-level memory. For each merchant, the model additionally generates a small set of \textit{Memory\_keywords} summarizing its distinctive products, services, combinations, or business format. Previously generated merchant names and core concepts are supplied to subsequent generations within the same category, encouraging alternative category-consistent concepts when substantial overlap occurs. This memory mechanism affects only textual construction and does not modify the structured merchant semantics.

After generation, duplicated merchant names are detected separately from semantic description construction. When a duplicate occurs, a dedicated renaming prompt takes the original merchant name and its generated description as input and produces a replacement name consistent with the existing business semantics. Only the merchant name is changed; the merchant description and structured attributes remain unchanged. The corresponding prompt is shown in Figure~\ref{fig:merchant_name_deduplication_prompt}.

\section{MS-FFSD Dataset Details}
\label{app:dataset_details}

\textbf{Dataset Fields.}
MS-FFSD organizes the released data into transaction, user, and merchant components. The transaction component records preserved transaction information together with enriched temporal and geographic semantics. The user and merchant components contain their respective structured semantic attributes and textual descriptions, linked to transactions through anonymized identifiers. Table~\ref{tab:msffsd_fields} provides the complete field definitions.

\begin{table}[ht]
\centering
\small
\caption{Field definitions of the released MS-FFSD dataset.}
\label{tab:msffsd_fields}

\setlength{\tabcolsep}{6pt}
\renewcommand{\arraystretch}{1.0}

\begin{tabular}{
    >{\raggedright\arraybackslash}m{0.24\linewidth}
    >{\raggedright\arraybackslash}m{0.68\linewidth}
}
\toprule

\rowcolor{my_purple}
\multicolumn{2}{c}{\textbf{Transaction}} \\
\midrule

Datetime
& Synthesized physical date and time constructed using IEEE-CIS-based temporal priors. \\

Source
& User identifier and foreign key to the user table. \\

Target
& Merchant identifier and foreign key to the merchant table. \\

Amount
& Original transaction amount preserved during semantic enrichment. \\

Location
& Original transaction-location identifier. \\

Type
& Original transaction-type identifier. \\

Transaction\_province
& Assigned physical province of the transaction. \\

Labels
& Transaction label indicating normal, fraudulent, or unlabeled status. \\

\midrule
\rowcolor{my_blue}
\multicolumn{2}{c}{\textbf{User}} \\
\midrule

Source
& User identifier and primary key. \\

Province
& Assigned home province of the user. \\

Gender
& User gender category. \\

Age\_group
& User age-group category. \\

Education
& User education-level category. \\

Occupation\_industry
& Industry associated with the user's occupation. \\

User\_description
& Textual description of the user's consumption behavior and semantic profile. \\

\midrule
\rowcolor{my_green}
\multicolumn{2}{c}{\textbf{Merchant}} \\
\midrule

Target
& Merchant identifier and primary key. \\

MCC\_level1
& Broad merchant consumption and business category. \\

MCC\_level2
& Fine-grained merchant category within MCC\_level1. \\

Merchant\_name
& Synthetic English merchant name. \\

Merchant\_description
& Textual description of the merchant's business characteristics and behavioral context. \\

\bottomrule
\end{tabular}
\end{table}

\textbf{User Semantic Attributes.} MS-FFSD enriches users with demographic, geographic, and occupational semantics. The province attribute follows the administrative divisions adopted in the dataset, while gender and education are sampled according to province-specific demographic distributions. Since transaction users are expected to have independent payment capability, individuals below 15 years old are excluded from age assignment. Age is represented as a categorical semantic attribute rather than a numerical value, with users grouped into Young Adults (15--44), Middle-aged Adults (45--59), and Older Adults ($\geq$60). For each gender, national age statistics are first aggregated into the three age groups to obtain the base proportions. The proportion of Older Adults is then adjusted for each province according to its relative 65+ population level, while the remaining probability is divided between Young Adults and Middle-aged Adults according to their gender-specific national proportions. The aging adjustment is left unbounded to retain regional differences. Occupation industry is represented using 19 industry categories and sampled according to province-specific employment distributions. The resulting categorical value spaces are summarized in Table~\ref{tab:user_semantic_categories}.

\begin{table}[ht]
\centering
\small
\caption{Categorical definitions of user-level semantic attributes in MS-FFSD.}
\label{tab:user_semantic_categories}

\setlength{\tabcolsep}{6pt}
\renewcommand{\arraystretch}{1.0}

\begin{tabular}{
    >{\centering\arraybackslash}m{0.27\linewidth}
    >{\raggedright\arraybackslash}m{0.65\linewidth}
}
\toprule

\rowcolor{my_blue}
\textbf{Attribute} &
\multicolumn{1}{c}{\textbf{Categories}} \\

\midrule

Gender
&
Male; Female. \\

\midrule

Age Group
&
Young Adults (15--44);
Middle-aged Adults (45--59);
Older Adults ($\geq$60). \\

\midrule

Education
&
Junior College and Above;
Senior Secondary School;
Junior Secondary School;
Primary School. \\

\midrule

Occupation Industry
&
Agriculture, Forestry, Animal Husbandry and Fishery;
Mining;
Manufacturing;
Production and Supply of Electricity, Heat, Gas and Water;
Construction;
Wholesale and Retail Trade;
Transport, Storage and Postal Services;
Accommodation and Food Services;
Information Transmission, Software and Information Technology Services;
Financial Services;
Real Estate;
Leasing and Business Services;
Scientific Research and Technical Services;
Water Conservancy, Environment and Public Facilities Management;
Resident Services, Repair and Other Services;
Education;
Health and Social Work;
Culture, Sports and Entertainment;
Public Administration, Social Security and Social Organizations. \\

\bottomrule
\end{tabular}
\end{table}

\textbf{Merchant Semantic Categories.} Merchant semantics are organized into a hierarchical category system to represent business activities at different levels of granularity. Each merchant is first associated with one of eight broad consumption categories and is then assigned an MCC Level 1 category describing its general business function, followed by a more specific MCC Level 2 category. This hierarchical design allows merchants within the same broad consumption domain to be further distinguished according to their business characteristics while maintaining a consistent category structure. The complete category hierarchy used in MS-FFSD is summarized in Table~\ref{tab:merchant_semantic_categories}.

\setlength{\tabcolsep}{6pt}
\renewcommand{\arraystretch}{1.0}

{
\small

\begin{longtable}{
    >{\centering\arraybackslash}m{0.27\linewidth}
    >{\raggedright\arraybackslash}m{0.65\linewidth}
}
\caption{Hierarchical definitions of merchant-level semantic categories in MS-FFSD.}
\label{tab:merchant_semantic_categories} \\

\toprule
\textbf{MCC Level 1} &
\multicolumn{1}{c}{\textbf{MCC Level 2}} \\
\midrule
\endfirsthead

\toprule
\textbf{MCC Level 1} &
\multicolumn{1}{c}{\textbf{MCC Level 2}} \\
\midrule
\endhead

\bottomrule
\endfoot


\rowcolor{my_green}
\multicolumn{2}{c}{Food, Tobacco and Liquor} \\
\midrule

Food, Tobacco and Liquor Retail
&
Convenience Store; Supermarket and Hypermarket; Fresh Produce Retail;
Specialty Food Retail; Tobacco, Liquor and Tea Retail. \\

\cmidrule(lr){1-2}

Food and Beverage Services
&
Fast Food and Snacks; Beverages, Bakery and Coffee; Cafeteria and Group Dining;
Full-Service Restaurant; Nightlife Bar and Dining;
Banquet and Large-Scale Catering. \\


\midrule
\rowcolor{my_green}
\multicolumn{2}{c}{Clothing and Footwear} \\
\midrule

Apparel and General Shopping
&
Apparel, Footwear and Bags; Department Store; Shopping Mall;
Outlet Shopping; Commercial Street Retail. \\


\midrule
\rowcolor{my_green}
\multicolumn{2}{c}{Housing} \\
\midrule

Housing Payments and Home Improvement Services
&
Utility Bill Payments; Property Management; Real Estate and Housing Agency;
Hardware and Building Materials; Home Furnishings and Textiles;
Building Decoration and Renovation Services;
Appliance, Furniture and Other Repair Services. \\


\midrule
\rowcolor{my_green}
\multicolumn{2}{c}{Household Equipment, Furnishings and Services} \\
\midrule

Household and Personal Goods Retail
&
Beauty and Personal Care; Mother and Baby Products and Children's Toys;
Office Supplies; Flowers and Green Plants; Pets and Pet Supplies;
Digital Electronics and Home Appliances. \\

\cmidrule(lr){1-2}

Local Lifestyle and Self-Service Sharing
&
Hair, Beauty and Nail Services; Bathing, Wellness and Health Services;
Housekeeping and Cleaning Services; Pet Hospital and Other Pet Services;
Digital and Entertainment Equipment Rental; Self-Service Sharing Services. \\


\midrule
\rowcolor{my_green}
\multicolumn{2}{c}{Transport and Communications} \\
\midrule

Local Commuting and On-Demand Travel
&
Urban Public Transport; Shared Two-Wheel Mobility Services;
On-Demand Ride-Hailing; Designated Driving and Parking Services. \\

\cmidrule(lr){1-2}

Long-Distance Travel Transport
&
Long-Distance Ground Travel; Air Travel Services;
Water Sightseeing Transport; Car Rental;
Expressway Toll and Service Areas. \\

\cmidrule(lr){1-2}

Vehicle Energy and Automotive Services
&
Fuel Supply Services; New Energy Charging and Battery Swap Services;
Vehicle Supplies, Repair and Maintenance. \\

\cmidrule(lr){1-2}

Telecommunications, Logistics and Digital Infrastructure Services
&
Telecom Payment and Operator Services; Information Search and Online Forums;
Internet Data Services; Software Development Services;
Express Delivery and Logistics Services; Basic Postal Services. \\


\midrule
\rowcolor{my_green}
\multicolumn{2}{c}{Education, Culture and Recreation} \\
\midrule

Education Payments and Training Services
&
Preschool Care Services; Primary and Secondary Education; Higher Education;
Youth Palace and Development Center; Other Education and Training. \\

\cmidrule(lr){1-2}

Offline Culture, Sports and Entertainment
&
KTV and Leisure Clubs; Board Games, Table Games and Gaming Cafes;
Fitness, Yoga and Dance; Cinema, Performances and Events;
Amusement Parks and Carnivals; Cultural and Sports Venues;
Books, Media, Arts and Musical Instruments; Outdoor Sports Equipment. \\

\cmidrule(lr){1-2}

Online Entertainment and Hobby Spending
&
Online Social Networking; Online Video, Audio and Reading;
Gaming; Live Streaming; Gaming Transactions and Related Services. \\

\cmidrule(lr){1-2}

Hotels, Scenic Attractions and Tourism Services
&
Hotels, Inns and Homestays; Scenic Attractions;
Travel Agencies and Tourism Services; Direct Tourism-Related Services;
Scenic Attraction Touring Services. \\


\midrule
\rowcolor{my_green}
\multicolumn{2}{c}{Health Care and Medical Services} \\
\midrule

Pharmaceutical, Medical Device and Health Retail
&
Pharmaceutical Sales; Nutrition and Health Products; Medical Device Sales;
Health and Assistive Therapy Equipment; Optical Store. \\

\cmidrule(lr){1-2}

General Clinical Care and Health Management
&
Public Primary Medical Care; Private Hospital;
Clinics and Independent Practitioners;
Medical Laboratories and Diagnostic Centers;
Health Examinations and Consultations. \\

\cmidrule(lr){1-2}

Specialized Consumer Medical and Care Services
&
Medical Aesthetics; Ophthalmic Medical Services;
Dental Medical Services; Online Medical Services;
Care Institution Services. \\


\midrule
\rowcolor{my_green}
\multicolumn{2}{c}{Miscellaneous Goods and Services} \\
\midrule

High-Value Goods and Professional Services
&
High-Value Jewelry and Accessories; Wedding and Photography Services;
Advertising, Exhibition and Printing Services; Legal Services;
Accounting and Financial Consulting Services; Recruitment Services;
Online Tools. \\

\end{longtable}

}

\textbf{Textual Descriptions.} The textual modality provides one entity-level description for each user and merchant. Financial transaction activity typically exhibits a long-tailed distribution, where a substantial fraction of users have only limited transaction histories~\citep{ref53}. MS-FFSD exhibits the same characteristic: as shown in Table~\ref{tab:textual_description_statistics}, 69.56\% of users have only one observed transaction. This imbalance in behavioral evidence motivates us to construct descriptions separately for single-transaction and multi-transaction users. For single-transaction users, a single observation is insufficient to support reliable aggregation of behavioral patterns, and forcing such aggregation may introduce unstable or weakly supported profile information. Their descriptions are therefore constructed deterministically from the available transaction and entity attributes. For multi-transaction users, repeated observations provide sufficient behavioral evidence to derive aggregated behavioral profiles and generate richer entity-level descriptions. Merchant descriptions are constructed at the entity level from merchant category semantics, transaction behavior, and associated user profiles.

Despite the different construction strategies, the resulting descriptions exhibit highly similar average lengths. Single-transaction and multi-transaction user descriptions contain 46.38 and 44.51 words on average, respectively, while the overall user average is 45.81 words. Merchant descriptions have a comparable average length of 46.94 words. This consistency indicates that neither the user-generation strategy nor the entity type introduces a substantial length discrepancy, reducing the likelihood that description length itself serves as a trivial shortcut or unintended signal in downstream modeling. The corresponding group sizes and description-length statistics are summarized in Table~\ref{tab:textual_description_statistics}.

\begin{table}[ht]
\centering
\small
\caption{Statistics of textual descriptions in MS-FFSD.}
\label{tab:textual_description_statistics}
\setlength{\tabcolsep}{4pt}
\renewcommand{\arraystretch}{1.0}

\begin{tabular}{
    >{\centering\arraybackslash}m{0.34\linewidth}
    >{\centering\arraybackslash}m{0.18\linewidth}
    >{\centering\arraybackslash}m{0.18\linewidth}
    >{\centering\arraybackslash}m{0.20\linewidth}
}
\toprule
\textbf{Description Group} &
\textbf{Number} &
\textbf{Proportion} &
\textbf{Avg. Length} \\
\midrule

\rowcolor{my_blue}
\multicolumn{4}{c}{User Descriptions} \\
\midrule

All Users
& 30,346
& --
& 45.81 words \\

Single-transaction Users
& 21,110
& 69.56\%
& 46.38 words \\

Multi-transaction Users
& 9,236
& 30.44\%
& 44.51 words \\

\midrule

\rowcolor{my_green}
\multicolumn{4}{c}{Merchant Descriptions} \\
\midrule

All Merchants
& 886
& --
& 46.94 words \\

\bottomrule
\end{tabular}
\end{table}

\section{Private Real-World Dataset Details}
\label{app:private_data}

Private-1 is a large-scale real-world financial transaction dataset containing diverse transaction patterns and substantial behavioral heterogeneity. Each transaction is associated with a label indicating whether it is normal, fraudulent, or unlabeled. The dataset records transaction order, interacting entities, transaction amount, and anonymized transaction context. Detailed definitions of the transaction attributes are provided in Table~\ref{tab:private1_dataset_detail}.

\begin{table}[ht]
\centering
\small
\caption{Descriptions of transaction fields in the Private-1 dataset.}
\label{tab:private1_dataset_detail}

\setlength{\tabcolsep}{6pt}
\renewcommand{\arraystretch}{1.0}

\begin{tabular}{
    >{\raggedright\arraybackslash}m{0.24\linewidth}
    >{\raggedright\arraybackslash}m{0.68\linewidth}
}
\toprule

\rowcolor{my_blue}
\textbf{Field} &
\multicolumn{1}{c}{\textbf{Description}} \\

\midrule
Time
& Global transaction order indicating the relative sequence of transactions \\

Source
& Anonymized identifier of the user \\

Target
& Anonymized identifier of the merchant \\

Amount
& Transaction amount \\

Location
& Anonymized location associated with the transaction \\

Type
& Anonymized transaction type \\

Label
& Transaction label indicating normal, fraudulent, or unlabeled status \\

\bottomrule
\end{tabular}
\end{table}

Private-2 is collected from a major commercial bank and consists of customer transaction records spanning four consecutive months from January to April. The records are organized into four monthly subsets in chronological order. Each transaction is associated with a label indicating whether it is normal, fraudulent, or unlabeled. Each record contains customer and merchant identifiers together with geographic, temporal, payment, and transaction attributes. Detailed definitions of all transaction fields are provided in Table~\ref{tab:private2_dataset_detail}.

\begin{table}[ht]
\centering
\small
\caption{Descriptions of transaction fields in the Private-2 dataset.}
\label{tab:private2_dataset_detail}

\setlength{\tabcolsep}{6pt}
\renewcommand{\arraystretch}{1.0}

\begin{tabular}{
    >{\raggedright\arraybackslash}m{0.24\linewidth}
    >{\raggedright\arraybackslash}m{0.68\linewidth}
}
\toprule

\rowcolor{my_green}
\textbf{Field} &
\multicolumn{1}{c}{\textbf{Description}} \\

\midrule
customer\_number
& Anonymized identifier of the transaction customer \\

merchant\_code
& Anonymized identifier of the merchant \\

receiving\_customer\_code
& Anonymized identifier of the merchant receiving the customer's transaction \\

card\_area
& Geographical region associated with the card \\

pre\_trade\_result
& Outcome of the previous transaction attempt \\

phone\_equal
& Whether the transaction phone matches the registered phone \\

trade\_time
& Time of the transaction \\

is\_common\_ip
& Whether the IP address is commonly used \\

e\_pay\_single\_limit
& Limit for a single electronic payment \\

e\_pay\_accumulate\_limit
& Cumulative electronic payment limit over time \\

trade\_amount
& Transaction amount \\

Label
& Transaction label indicating normal, fraudulent, or unlabeled status \\

\bottomrule
\end{tabular}
\end{table}

The two private datasets differ substantially in both temporal structure and user transaction-frequency distribution. First, their original temporal fields encode transaction order differently. In Private-1, the temporal index is globally unique and strictly increasing, providing a complete ordering of all transactions. In Private-2, the temporal index is non-decreasing but may contain duplicate values, meaning that multiple transactions can share the same index and their relative order cannot be further distinguished. Private-2 additionally provides explicit monthly boundaries. The Temporal Agent therefore adapts its timestamp construction to the available ordering information: Private-1 preserves the unique global order, whereas transactions sharing the same index in Private-2 are handled within the same temporal bucket without imposing an unsupported intra-bucket ordering.

For graph construction and behavioral feature extraction, the dataset-specific schema is mapped to the unified representation used in Appendix~\ref{app:graph_details}. In Private-2, customer\_number and merchant\_code are used as the user and merchant identifiers, respectively. Card\_area is used as the geographic attribute corresponding to Location, while the dataset-specific transaction-context attributes are encoded into a unified Type representation. The resulting Location and Type representations are then used to construct the corresponding structural relations and the same 12-dimensional behavioral features as in the other datasets.

The two datasets also exhibit markedly different user activity distributions. Private-1 shows a pronounced long-tailed pattern, with most users associated with only a single observed transaction, whereas single-transaction users constitute only a minority in Private-2. This difference reflects substantially different levels of transaction-history sparsity and user behavioral coverage, and may also arise from different data collection or preprocessing regimes. These differences provide distinct conditions for evaluating the semantic enrichment framework. The framework is applied to both datasets without relying on a single temporal representation or user-activity distribution, and the resulting multimodal data improve downstream fraud detection in both settings. The consistent gains across these different data conditions therefore provide evidence for the generalizability of the proposed semantic enrichment framework.

\section{Graph-based Fraud Detection Details}
\label{app:graph_details}

This section provides implementation details for graph-based fraud detection, covering graph construction, semantic text construction, feature encoding, and experimental settings.

\subsection{Graph Construction}

We construct an entity graph in which users and merchants are represented as nodes, rather than treating individual transactions as graph nodes. This design is motivated by the granularity of the generated semantics. User descriptions characterize demographic profiles, consumption preferences, and behavioral patterns over the observation period, while merchant descriptions capture business categories, service characteristics, and customer profiles. These semantics therefore correspond naturally to persistent entities rather than individual transactions.

A transaction-level graph would repeatedly attach the same entity descriptions to many transaction nodes and weaken the alignment between node identity and semantic information. More importantly, for transaction-time fraud detection, descriptions summarized from the full observation period would expose future behavioral information to earlier transactions. Avoiding such leakage would require dynamically updating entity descriptions at each transaction time, substantially increasing both generation and encoding costs. Under our user-level fraud identification setting after the complete observation period, the entity graph provides a more consistent alignment among prediction targets, node labels, structured attributes, and textual descriptions.

\textbf{Nodes and Labels.}
The graph contains two types of entities: users and merchants. Users serve as the target nodes for fraud detection, while merchants are retained as background nodes that provide transaction connectivity and merchant-side semantic context. Following the use of background nodes in financial graph anomaly detection~\citep{ref36}, merchant nodes participate in message passing but are excluded from the supervised prediction objective.

User labels are derived from their associated transaction labels using a priority rule. A user is labeled as fraudulent (\(y=1\)) if they have initiated at least one fraudulent transaction. Among the remaining users, those associated with unlabeled transactions are assigned \(y=2\), while all others are labeled as normal (\(y=0\)). All merchant nodes are assigned the background label \(y=3\). Only labeled users with \(y\in\{0,1\}\) participate in supervised training, validation, and testing, whereas unlabeled users and background merchants remain in the graph to preserve structural connectivity and support message passing. This node-level distribution differs substantially from the transaction-level label distribution in Table~\ref{tab:sffsd_statistics}, since any user with at least one fraudulent transaction is labeled fraudulent, whereas users containing unlabeled but no fraudulent transactions are excluded from supervised evaluation.

\textbf{Edge Construction.}
We construct five types of entity relations from the original transaction records. The original User--Merchant transaction edges are preserved directly, while all induced structural relations are scored using inverse-document-frequency (IDF) weighting and sparsified with Mutual Top-\(K\). IDF weighting reduces the contribution of highly frequent entities or attributes that provide weak discriminative information and assigns larger scores to rarer shared relations. When a node pair is supported by multiple shared entities, attributes, or repeated co-occurrences, their contributions are accumulated into a candidate-edge score. Mutual Top-\(K\) then retains an edge only when each endpoint ranks the other among its \(K\) highest-scoring candidate neighbors. This procedure filters asymmetric weak associations, limits excessive connectivity around high-degree nodes, and preserves relatively stable bidirectional relations.

The five relation types are constructed as follows:

\begin{enumerate}[leftmargin=15pt,topsep=0pt]
\setlength\itemsep{0em}
    \item U--M: User--Merchant Transaction Relation. A user and a merchant are connected if at least one transaction occurs between them. These edges directly preserve the observed user--merchant interactions and are not subject to IDF weighting or Mutual Top-\(K\) sparsification.

    \item U--W--U: Local Temporal Co-occurrence Relation. For each merchant, its associated transactions are ordered by time. A local window of length temporal\_window is constructed around each transaction, and distinct users appearing within the same window form candidate user--user relations. These candidates are subsequently scored and sparsified using IDF and Mutual Top-\(K\).

    \item U--A--U: Temporal Attribute-Matching Relation. All transactions are globally ordered by time, and only adjacent transactions are compared. When two adjacent transactions originate from different users and share either the same Location or the same Type, their users form a candidate relation. IDF weighting and Mutual Top-\(K\) are then applied.

    \item U--M--U: Shared-Merchant Relation. Two users form a candidate relation when they have transacted with the same merchant. Unlike the local temporal relation above, this relation captures shared merchants over the full observation period. Candidate edges are filtered using IDF and Mutual Top-\(K\).

    \item M--U--M: Shared-User Relation. Merchants connected to the same user form candidate merchant--merchant relations. This relation is structurally symmetric to the shared-merchant relation and captures merchants linked through common users. The same IDF and Mutual Top-\(K\) procedure is applied.
\end{enumerate}

After sparsification, all five relation types are merged into a single homogeneous graph. The final graph does not expose relation types or IDF scores to downstream detectors, and all retained edges are treated as unweighted connections. IDF weighting and Mutual Top-\(K\) are therefore used only during graph construction to select informative structural relations.

\textbf{Original Node Features.}
For each user or merchant node, the associated transaction records are aggregated into a 12-dimensional behavioral representation. The features characterize four aspects of transaction behavior. First, transaction activity is described by the total number of transactions and the accumulated transaction amount. Second, transaction magnitude and dispersion are captured through the maximum amount, standard deviation, three quartiles of the amount distribution, and skewness, which together reflect typical spending levels, variability, and the presence of long-tailed or unusually large transactions. Third, behavioral concentration is measured by the proportions of transactions occurring at the most frequently used location and transaction type. Finally, the diversity of locations and transaction types is quantified using entropy-based measures, with larger values indicating more dispersed geographic activity or more diverse transaction behaviors. These aggregated statistics form the original node representation before projection and semantic fusion.

\subsection{Semantic Text Construction}

Semantic attributes are converted into natural-language sequences before embedding. Rather than encoding each semantic field independently, we serialize all fields of an entity into a unified sequence before embedding. This design allows the embedding model to capture semantic relationships across attributes and provides richer context for fields that are less informative in isolation. It also produces a fixed-dimensional representation regardless of the number of fields, avoiding field-specific dimensional allocation and simplifying the incorporation of additional attributes. Moreover, encoding a complete semantic description better matches the natural-language input form of text embedding models than encoding isolated and often semantically sparse field fragments.

\begin{figure}[htb]
    \centering

    \begin{subfigure}{1.0\linewidth}
        \centering
        \includegraphics[width=\linewidth]{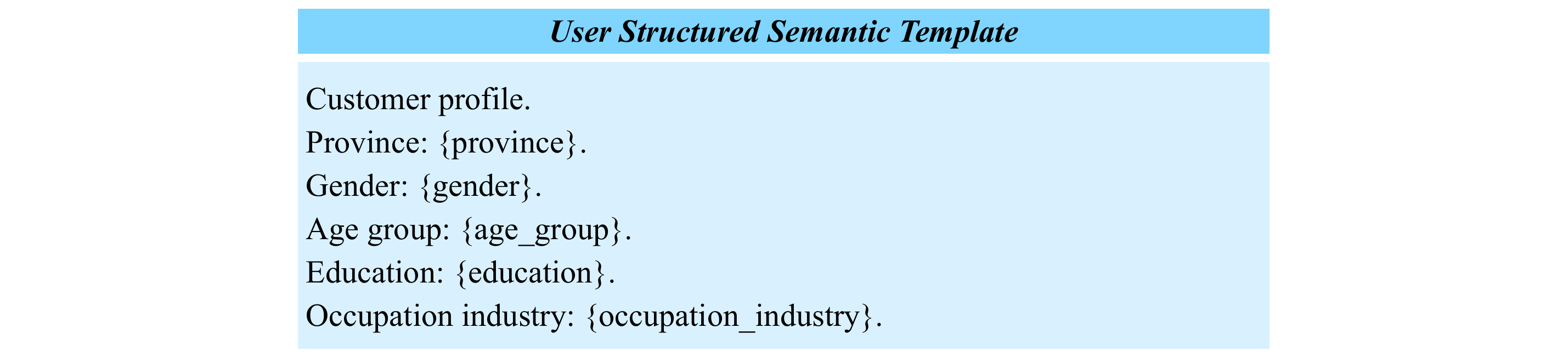}
        \captionsetup{width=0.6\linewidth}
        \caption{User structured semantic template containing user demographic and geographic information.}
        \label{fig:user_template_structured}
    \end{subfigure}

    \vspace{6pt}

    \begin{subfigure}{1.0\linewidth}
        \centering
        \includegraphics[width=\linewidth]{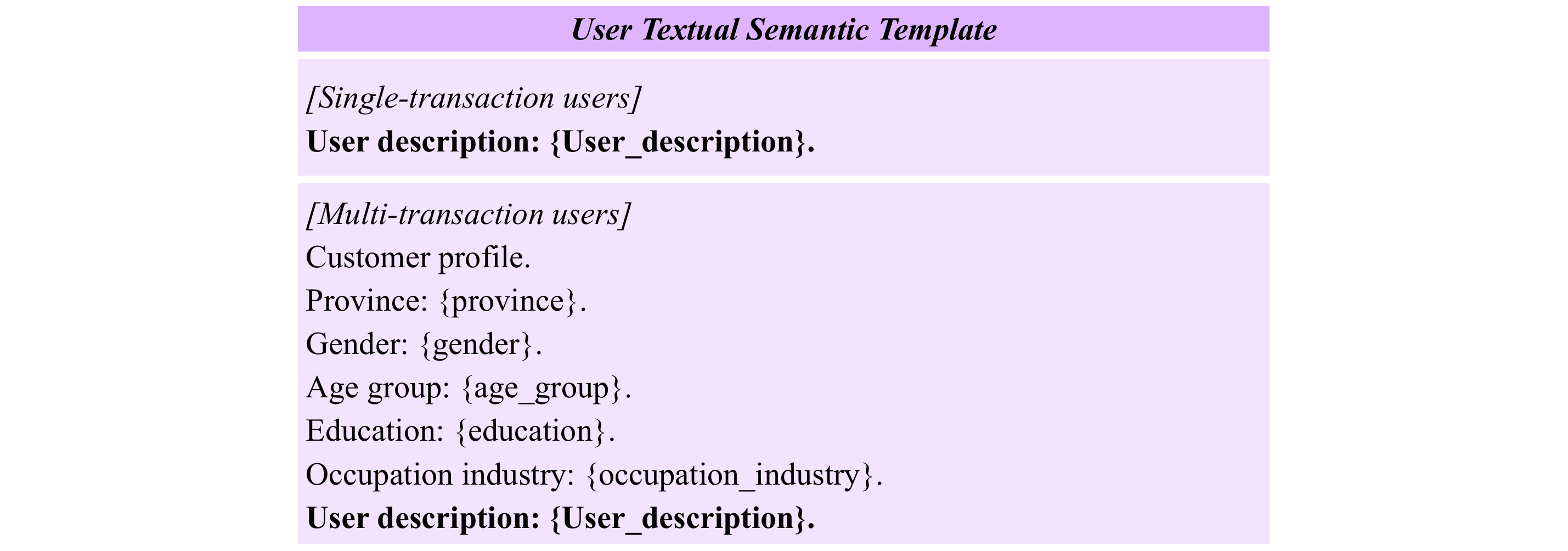}
        \captionsetup{width=0.6\linewidth}
        \caption{User textual semantic template containing user profile information and entity-level behavioral descriptions.}
        \label{fig:user_template_textual}
    \end{subfigure}

    \caption{Semantic templates for user.}
    \label{fig:user_semantic_templates}
\end{figure}

For user nodes, the structured semantic template in Figure~\ref{fig:user_semantic_templates}(\subref{fig:user_template_structured}) serializes the generated demographic and geographic information into a unified user profile. When using textual semantics, user inputs are serialized according to the description-generation strategy. For single-transaction users, the deterministic description already verbalizes the structured user attributes, and only the description is therefore retained. For multi-transaction users, whose LLM-generated descriptions explicitly exclude demographic information, the structured user profile is prepended to the behavioral description. This ensures that structured profile information is represented once in both cases while avoiding redundant semantic content, as shown in Figure~\ref{fig:user_semantic_templates}(\subref{fig:user_template_textual}). This allows explicit profile attributes and higher-level behavioral characteristics inferred from the observed transaction history to be jointly represented within the same semantic sequence.

\begin{figure}[htb]
    \centering

    \begin{subfigure}{1.0\linewidth}
        \centering
        \includegraphics[width=\linewidth]{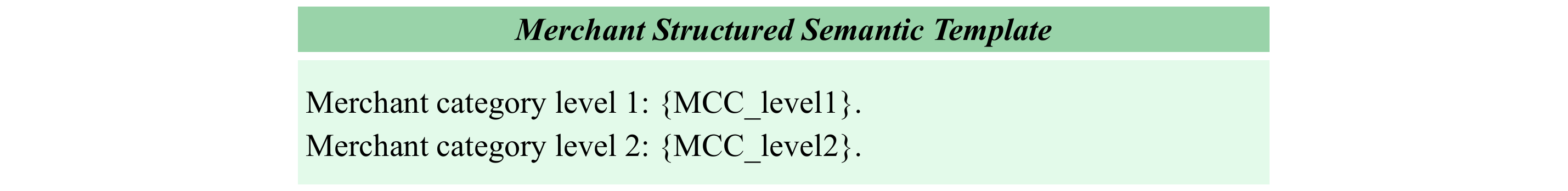}
        \captionsetup{width=0.6\linewidth}
        \caption{Merchant structured semantic template containing hierarchical business category information.}
        \label{fig:merchant_template_structured}
    \end{subfigure}

    \vspace{6pt}

    \begin{subfigure}{1.0\linewidth}
        \centering
        \includegraphics[width=\linewidth]{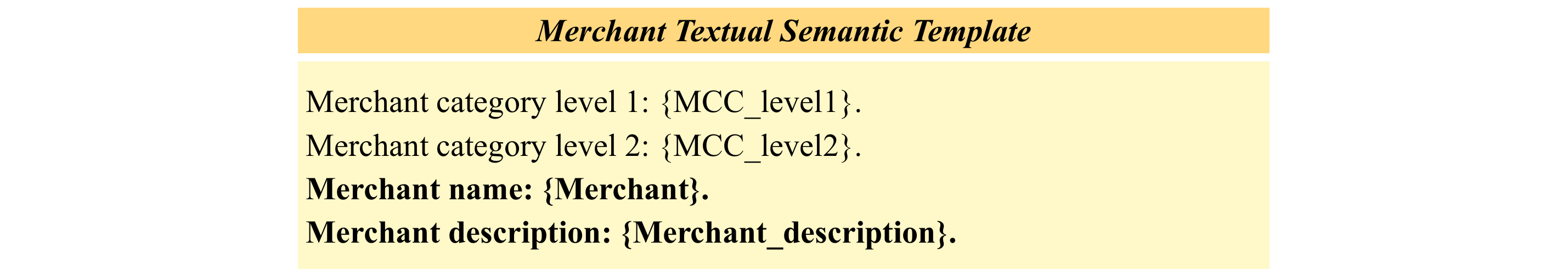}
        \captionsetup{width=0.6\linewidth}
        \caption{Merchant textual semantic template containing business category information, merchant name, and entity business descriptions.}
        \label{fig:merchant_template_textual}
    \end{subfigure}

    \caption{Semantic templates for merchant.}
    \label{fig:merchant_semantic_templates}
\end{figure}

For merchant nodes, the structured semantic template in Figure~\ref{fig:merchant_semantic_templates}(\subref{fig:merchant_template_structured}) organizes the generated hierarchical business-category information into a unified merchant representation. When using textual semantics, the same category information is retained while merchant identity and the entity business description are further incorporated, as illustrated in Figure~\ref{fig:merchant_semantic_templates}(\subref{fig:merchant_template_textual}). The resulting sequence jointly captures business category, merchant identity, and business context.

Importantly, textual semantic inputs are constructed within the same semantic encoding pipeline rather than through an additional feature branch. For user nodes, structured profile information is included only once, avoiding repetition when it is already verbalized in the deterministic description of single-transaction users. Thus, structured and textual semantics share the same semantic encoding pipeline, with their difference arising from the semantic content included in the input sequence.

\subsection{Feature Encoding}

The original features and semantic information are processed through separate encoding branches before being provided to the graph-based detectors. For the original branch, the 12-dimensional node features are standardized using statistics computed from the training set and mapped to a 128-dimensional representation through a node-type-specific linear layer followed by LayerNorm and ReLU. User and merchant nodes use separate projection and normalization parameters to accommodate their different feature distributions.

For the semantic branch, the serialized entity text constructed above is encoded using Qwen3-Embedding-8B~\citep{ref30}. Leveraging its Matryoshka representation capability, we directly truncate the original 4096-dimensional output to 128 dimensions during encoding and apply L2 normalization, without additional dimensionality reduction such as PCA. These semantic embeddings are precomputed offline and kept fixed during graph-model training, without updating the parameters of Qwen3-Embedding-8B. The fixed embeddings are subsequently processed through trainable node-type-specific linear projections, followed by LayerNorm and ReLU.

Under the original features setting, the projected 128-dimensional original representation is directly provided to the graph model. When structured or textual semantics are introduced, the two 128-dimensional branch representations are first concatenated and then mapped back to 128 dimensions through a node-type-specific fusion projection followed by LayerNorm and ReLU. Consequently, all three settings provide the graph backbone with the same 128-dimensional node input, keeping the backbone input dimensionality fixed across semantic configurations.

\subsection{Experimental Settings}

\textbf{Implementation Details.}
For all experiments, labeled user nodes are randomly divided into training, validation, and test sets at a ratio of 6:2:2, and identical splits are used across models and semantic settings for fair comparison. Optimization is performed using Adam with a learning rate of 0.001 and a weight decay of 0.001. Training is capped at 300 epochs with an early-stopping patience of 50 epochs based on validation loss, and the checkpoint with the lowest validation loss is used for test evaluation. Model-specific configurations follow the officially recommended settings of the corresponding graph-based detectors. Each experiment is repeated over five random seeds, and the mean performance is reported. All experiments are conducted on a server equipped with four 32GB NVIDIA Tesla V100 GPUs.

The larger training portion provides sufficient supervision for learning how the introduced semantics interact with the original features, while retaining separate validation and test sets for model selection and reliable evaluation. The 300-epoch upper bound provides sufficient optimization time for models with different convergence behaviors, while early stopping with a patience of 50 epochs prevents unnecessary prolonged training and reduces overfitting.

\textbf{Evaluation Metrics.}
We evaluate fraud detection performance using AUC, AP, and F1-macro, which characterize model effectiveness from different aspects. AUC measures the overall ranking ability between fraudulent and normal samples across different decision thresholds. AP focuses on the ranking quality of fraudulent samples and is particularly informative under class imbalance. F1-macro evaluates classification performance by computing F1 scores for both classes and assigning them equal importance, thereby reducing the influence of the majority class. Together, these three metrics provide a comprehensive evaluation of model performance.

\section{LLM-based Semantic Reasoning Details}
\label{app:llm_reasoning}

\subsection{Reasoning Setup}

We conduct zero-shot semantic reasoning using DeepSeek-V3~\citep{ref52}. To isolate the effect of semantic enrichment, the same model configuration and fixed reasoning prompt are used across all enrichment levels, with only the information provided in the input changing accordingly. Ground-truth fraud labels are excluded from the input, so the model must form its assessment solely from the observed behavioral and semantic evidence.

The reasoning task examines how enriched semantics change transaction behavior interpretation. The prompt asks the model to identify salient evidence, determine whether available semantics contextualize each observation, and produce an evidence-grounded assessment. This avoids treating semantic consistency as evidence of normality while retaining unexplained abnormalities.

The reasoning input follows the same three progressive information settings used in graph-based fraud detection. Under \textit{Original Features}, the input contains the target user's behavioral features and transaction-derived information associated with its connected merchants, without generated semantic attributes or textual descriptions. Under \textit{Structured Semantics}, the generated user profile and merchant business-category information are additionally provided. Under \textit{Textual Semantics}, the structured information is retained and further enriched with the user behavioral description and merchant identity and business descriptions. Across the three settings, the underlying target user and associated merchants remain unchanged, with only the available semantic information progressively expanded. This enables direct comparison of how additional semantic context affects the interpretation of the same observed behavior. Unlike graph models, which consume projected numerical representations and semantic embeddings, the LLM directly receives the corresponding values and semantic information in textual form.

\subsection{Reasoning Prompt and Output Protocol}

\begin{figure}[ht]
    \centering
    \includegraphics[width=\linewidth]{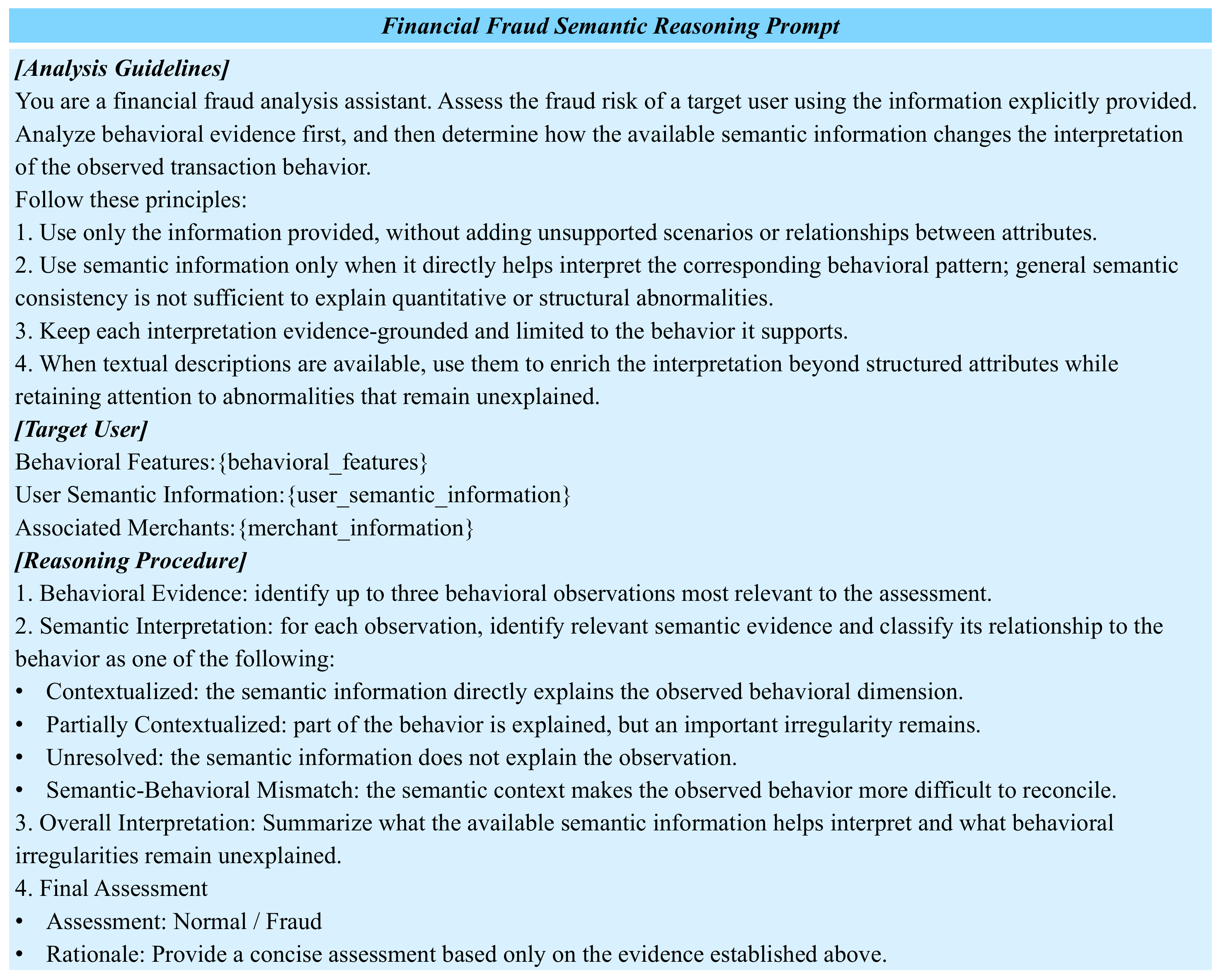}
    \captionsetup{width=0.9\linewidth}
    \caption{Prompt used for LLM-based semantic reasoning across all enrichment settings.}
    \label{fig:llm_reasoning_prompt}
\end{figure}

A fixed reasoning prompt is applied across all cases and enrichment levels, as shown in Figure~\ref{fig:llm_reasoning_prompt}. The prompt consists of three components: analysis guidelines, target-user information, and a structured reasoning procedure. The analysis guidelines restrict the model to explicitly provided evidence and require semantic explanations to correspond to the specific behavioral patterns being interpreted. Textual descriptions may provide additional context beyond structured attributes, while abnormalities that remain unexplained are retained in the reasoning process.

For each target user, the model first identifies up to three behavioral observations relevant to the assessment. It then associates each observation with available semantic evidence and classifies their relationship as contextualized, partially contextualized, unresolved, or semantic--behavioral mismatch. The model subsequently summarizes the overall interpretation and returns a final normal or fraud assessment with a concise evidence-grounded rationale.

This structured output enables comparison across enrichment levels at both the behavioral-interpretation and final-assessment levels, allowing us to examine how structured and textual semantics progressively influence contextual reasoning under the same transaction evidence.

\subsection{Evaluation Scope and Case Selection}

The LLM experiment is designed as a qualitative zero-shot reasoning analysis rather than a predictive benchmark. It examines whether progressively enriched semantic information supports more contextual interpretation of the same transaction behavior, rather than evaluating the LLM as a supervised fraud detector. We therefore do not report classification metrics for this experiment, but focus on how the reasoning process changes as structured and textual semantics are progressively introduced.

Illustrative cases for qualitative analysis are selected according to the following criteria:

\begin{itemize}[leftmargin=15pt,topsep=0pt] \setlength\itemsep{0em}
    \item \textbf{Cross-setting comparability.} The same target user must have complete reasoning outputs under all three enrichment settings, so that changes in the reasoning process can be attributed to the progressively introduced semantic information rather than differences in the analyzed case. This ensures that the comparison is conducted under a consistent behavioral context.

    \item \textbf{Reasoning contrast across settings.} The selected case should allow meaningful comparison of the reasoning process as semantic information is progressively introduced. Such differences may appear in the identified evidence, behavioral interpretation, or the connection between observed behavior and semantic context.

    \item \textbf{Diversity of semantic effects.} The selected cases should reflect different effects of semantic enrichment, including cases where additional context helps explain ambiguous behavior and cases where important abnormalities remain only partially contextualized or unresolved.
\end{itemize}

\end{document}